\documentclass[biblatex,cm]{glossa}

\let\B\relax %
\let\T\relax

\usepackage[utf8]{inputenc}
\usepackage{amsmath}
\usepackage{graphicx}
\usepackage{amssymb} 
\usepackage{url}
\usepackage[above,below]{placeins}  %
\usepackage[backend=biber]{biblatex}  %
\usepackage{randtext}    %

\pdfauthor{Karjus et al.}
\pdftitle{Challenges in detecting evolutionary forces in language change using diachronic corpora}
\pdfkeywords{language evolution, language change, selection, drift, corpus-based, temporal binning}

\title[Challenges in detecting evolutionary forces in language change]{Challenges in detecting evolutionary forces in language change using diachronic corpora}

\author[Karjus et al.]%
{              %
  \spauthor{Andres Karjus\\ 
  \institute{Centre for Language Evolution, University of Edinburgh, UK}
  \small{Corresponding author (\randomize{a.karjus}@sms.ed.ac.uk)
  }
}
  \AND
  \spauthor{Richard A. Blythe\\
  \institute{School of Physics and Astronomy; Centre for Language Evolution, University of Edinburgh, UK}
  }%
  \AND
  \spauthor{Simon Kirby\\
  	\institute{Centre for Language Evolution, University of Edinburgh, UK}
  }
  \AND
  \spauthor{Kenny Smith\\
  \institute{Centre for Language Evolution, University of Edinburgh, UK}
}
}

\begin{document}

\sffamily
\maketitle

\begin{abstract}
	Newberry~et al.\ (Detecting evolutionary forces in language change, \textit{Nature} 551, 2017) tackle an important but difficult problem in linguistics, the testing of selective theories of language change against a null model of drift. 
	Having applied a test from population genetics (the Frequency Increment Test) to a number of relevant examples, they suggest stochasticity has a previously under-appreciated role in language evolution.
	We replicate their results and find that while the overall observation holds, results produced by this approach on individual time series can be sensitive to how the corpus is organized into temporal segments (binning).  
	Furthermore, we use a large set of simulations in conjunction with binning to systematically explore the range of applicability of the Frequency Increment Test.
	We conclude that care should be exercised with interpreting results of tests like the Frequency Increment Test on individual 
	series, given the researcher degrees of freedom available when applying the test to corpus data, and fundamental differences between genetic and linguistic data. 
	Our findings have implications for selection testing and temporal binning in general, as well as demonstrating the usefulness of simulations for evaluating methods newly introduced to the field. 
\end{abstract}


\rmfamily

\section{Introduction}

All natural languages change over time. The way each new generation of speakers pronounces their words is subtly different from their parents, new words replace old ones, marginal grammatical paradigms become the norm, and norms dissolve.
Many authors have suggested that language change, like other evolutionary processes, involves both directed selection as well as stochastic drift \parencite{sapir_language._1921,jespersen_language_1922,andersen_structure_1990,mcmahon_understanding_1994,croft_explaining_2000,baxter_utterance_2006,van_de_velde_degeneracy_2014,steels_evolutionary_2018}.  
Systematically quantifying the relative contribution of these two processes --- particularly with reference to individual time series --- is an open problem. 

There are a number of ways in which selective biases may influence language change. 
For example various cognitive biases have been postulated as important in the evolution of language \parencite{kirby_cumulative_2008,fay_interactive_2010,smith_linguistic_2013,tamariz_cultural_2014,enfield_transmission_2014,croft_explaining_2000,haspelmath_optimality_1999} and one might therefore expect to see manifestations of these in instances of language change.  
Selective advantage stemming from sociolinguistic prestige of (the users of) competing variants has been shown to play a considerable role in change, both via competition between forms within the language community as well as borrowing from other languages \parencite{labov_principles_2011,hernandez-campoy_handbook_2012}.
A foreign or novel variant may also be selected for by virtue of filling a lexical or morphosyntactic gap \parencite{mcmahon_understanding_1994,trask_historical_1996}.
The form of a variant alone may convey a selective advantage. 
For example, it has been observed that, all other things being equal, speakers prefer shorter forms that take less effort to utter \parencite{zipf_human_1949,kanwal_zipfs_2017} and limited iconicity can be advantageous \parencite{dingemanse_arbitrariness_2015}.
Various usage and acquisition properties have been shown to be predictors of success \parencite{kershaw_modelling_2016,calude_modelling_2017,grieve_mapping_2018,monaghan_cognitive_2019}.
There is also evidence that certain phonetic changes are more likely than others, due to the articulatory and acoustic properties of human speech sounds \parencite{ohala_origin_1983,baxter_utterance_2006}.
In certain circumstances there may be even qualitative evidence of directed selection, such as knowledge of previous activities of some authoritative language planning body, prescriptive grammars, or other exogenous forces \parencite{ghanbarnejad_extracting_2014,daoust_language_2017,rubin_language_1977,anderwald_variable_2012}.

It is a reasonable hypothesis that, given adequately large and representative samples of language use over time (i.e., corpora), signatures of selection should be inferable from the usage data alone. This idea has recently been explored in a number of works \parencite{sindi_culturomics_2016,reali_words_2010,blythe_neutral_2012,bentley_random_2008,amato_dynamics_2018,hahn_drift_2003}, and has been also applied to domains of cumulative cultural evolution beyond language \parencite{kandler_inferring_2017,kandler_analysing_2019}.
One of the more ambitious attempts is that of \textcite{newberry_detecting_2017}, who employ a standard method borrowed from the field of population genetics, which also deals with the inference of selection in a population and the assessment of drift in evolution (we will henceforth refer to this work as `Newberry~et~al.'; \textcite{ahern_evolutionary_2016} is an earlier version of the paper). They use the Frequency Increment Test \parencite{feder_identifying_2014}, or FIT for short, and make an explicit connection with the Wright-Fisher model \parencite{wright_evolution_1931,ewens_mathematical_2004} of neutral stochastic drift (not unlike a previous similar contribution \parencite{sindi_culturomics_2016}).

Newberry~et~al.\ consider three grammatical changes in the English language. Their main focus is the (ir)regularization of past-tense verbs (e.g. the change from irregular \textit{snuck} to regular \textit{sneaked}), a topic that has been of some interest \parencite{lieberman_quantifying_2007,cuskley_internal_2014,gray_english_2018}.
They also investigate the change in periphrastic `do' (\textit{say not that!} becoming \textit{don't say that!}), the evolution of verbal negation (from the Old English pre-verbal to the Early Modern English post-verbal), and possible phonological neighborhood effects (which we will not discuss here).
They use data from the Corpus of Historical American English \parencite{davies_corpus_2010} and the Penn Parsed Corpora of Historical English \parencite{kroch_penn-helsinki_2000}.
Their method consists of calculating the relative frequencies of alternative forms in a corpus (e.g., the relative frequency of the irregular past tense form \textit{snuck} against that of the regular \textit{sneaked}), placing the count data into variable-length temporal bins, and running the FIT on the resulting time series. Ultimately, the test yields a $p$-value under the null hypothesis of change by drift alone.
They also infer the ``effective population size" of the verbs and show that the strength of drift (in a subset of verbs with a FIT $p>0.2$) correlates inversely with corpus frequencies, echoing the analogous observation about small populations in genetics.

The FIT points towards selection being operative in some cases, while labelling others (in fact, most changes in past-tense forms) as changes stemming from drift. 
In this work, we replicate this analysis (using Newberry~et~al.'s original code; see the Data Availability section in the end). 
We highlight an important methodological issue that arises when applying the FIT to linguistic data and which should be taken into account in future applications of the FIT (and similar tests) to identify cases of selection from linguistic corpora. 
The key issue lies in the construction of the time series via binning counts (e.g. from a corpus), and the application of the test in question to such time series, but we also draw attention to issues more specific to diachronic language data. 
While the FIT may be an appropriate test in some cases, we show that an incautious application of the FIT to linguistic data can end up incorrectly identifying cases of drift as cases of selection, and missing subjectively clear cases of selection.

While the approach of applying a test of selection to corpus-based time series shows promise as a method of linguistic analysis, we believe these issues deserve further investigation. We briefly explain the technical aspects of temporal binning and the FIT in the next subsections.

\subsection{Linguistic corpora and data binning}\label{section:corporabinning}

In quantitative research on language dynamics, words and grammatical constructions are often equated with alleles \parencite{reali_words_2010}. This analogy is motivated by the observation that a given `underlying form' may have two or more (near-) synonymous actualizations or `surface forms' (e.g. as in the \textit{sneaked}--\textit{snuck} case which are both actualizations of \textit{sneak}.\textsc{past}).
Word variants are not quite like alleles though. Organisms inherit genetic material from their parents, and one can (in principle) test for the presence of a particular allele in each individual in the population over time. In the context of language use, the notions of parents, offspring and generations are more diffuse than they are in genetics. What is done in practice when analysing time series is to construct an artificial `generation' by collecting together all instances of the word variants under consideration that fall within a specific time window (or `bin'). Particularly troublesome is that fact that a given lexeme may not occur in a given corpus in a particular period of time, which means having to widen the bin to obtain a meaningful frequency.  Such absences may occur simply because of the finite size of the sample: any corpus is in the end just a sample from a population of utterances. The smaller the corpus, the smaller the chance a lexeme has to occur. It may also be because people talked and wrote about other topics in that time window, which did not require the use of this particular sense. A corpus may be large, but not well balanced, in the sense that it does not cover all the relevant genres or topics of the time (a critique recently directed by \textcite{pechenick_characterizing_2015} at another widely used corpus, the Google Books N-grams dataset).

To understand the issue of binning (or temporal segmentation) in more detail, let us consider for a moment a fictional corpus of a daily newspaper, spanning two centuries. 
Our goal is to count the occurrences of two competing spelling forms of a word and operationalise these as relative frequencies in a time series. 
The smallest possible temporal sample would consist of the text that makes up one daily issue of the paper (yielding a fine grained time series of about $n=73000$ data points). One could also aggregate (bin) all the texts from one month ($n=2400$), year ($n=200$), decade ($n=20$) or century ($n=2$). 
However, there is no single ideal way to bin the data. A century, with only two data points, may be too large a chunk, as it may miss processes taking place in between --- and it is difficult to infer anything about the dynamics of the change from two data points. 
A day is likely too small a sample, since the word (in either spelling) might not occur every day, unless it is a particularly commonly used one.

In corpus-based language research either years or decades therefore seem the most commonly used bins. Regardless, a decision has to be made regarding how to bin corpus data; our point here is to show that this decision (which potentially constitutes an additional researcher degree of freedom, since different binning decisions may yield different results) influences the outcome of analyses which use tests like the FIT to identify selection.

\subsection{The Frequency Increment Test}\label{section:fittest}

The FIT \parencite{feder_identifying_2014} belongs to a family of methods conceived to detect selection in time series genetic data, with intended application to population genetics experiments and historic DNA samples. All of them boil down to looking for certain patterns in time series of allele frequencies \parencite{iranmehr_clear_2017,nishino_detecting_2013,terhorst_multi-locus_2015,schraiber_bayesian_2016} (see also \parencite{malaspinas_methods_2016} for a review).
Such approaches rely on the presumption that a change driven by selection would look different, or leave different `signatures', from a change happening due to stochastic drift.

The FIT works as follows. Relative frequencies in the range $(0,1)$ are transformed into frequency increments $Y$ according to
\begin{equation}
Y_i = (v_i - v_{i-1}) / \sqrt{ 2v_{i-1}(1-v_{i-1})(t_i-t_{i-1}) }
\end{equation}
where $v_i$ is the relative frequency of a variant at a measurement time $t_i$. The rationale behind this rescaling is that, under neutral evolution, the mean increment $v_i-v_{i-1}$ (i.e. the change in frequency of $v_i$ from time $t_{i-1}$ to time $t_i$) is zero, and its variance is proportional to 
\begin{equation}
v_{i-1}(1-v_{i-1})(t_i-t_{i-1}) \;,
\end{equation} 
i.e. the expected variance under drift is large when we are looking at the changes in frequency between two widely separated time points (i.e. $t_{i-1}$ and $t_i$ are far apart) or when values of $v_{i}$ are close to 0.5 (i.e. changes in frequency driven by drift will tend to be small when the variant is very rare and $v_{i}$ is close to 0, or very common and $v_{i}$ is close to 1).

The FIT relies on the Gaussian approximation of the Wright-Fisher diffusion process.
When the variant frequency $v_i$ is not too close to either of the boundary values $0$ or $1$ and the time between successive measurements is sufficiently small, the random variables $Y_i$ can be approximated as having a normal distribution with a mean of zero and a variance that is inversely proportional to an effective population size (which is taken to be constant over time). 
Thus a test under the null hypothesis of drift amounts to a test of how likely the transformed increments $Y_i$ are under the assumption that they are drawn from a normal distribution with a mean of zero, as would be the case under drift: this can be evaluated using a one-sample $t$-test test under the assumption of normally-distributed increments with a zero mean and equal variance.

In this context, a failure to reject the null indicates a failure to reject the hypothesis of drift. On the other hand, if the null hypothesis is rejected, than the changes may be due to some non-neutral process. 
In this work, we check for the normality assumption using the Shapiro-Wilk test. Homoscedasticity (the assumption that the underlying distributions have equal variances) is less straightforward; we explore its relevance in the Supplementary Appendix.

The authors of the Frequency Increment Test \parencite{feder_identifying_2014} note that its power increases with the number of sampled time points, but also that it has low power in cases of both very weak (near-drift) and very strong selection coefficients.
The latter leads to a situation where fixation to a variant happens swiftly within the sampling interval (the range of the time series), making the rest of the time series uninformative. The frequencies should also be far from absorbing boundaries (i.e., situations where one variant is at (or near) 0\% and the other at 100\% of the population), which might pose a particular problem in corpus-based time series analysis: since linguistic change is (classically) believed to follow an S-shaped trajectory \parencite{blythe_s-curves_2012}, a change which takes place near the start or end of a given corpus would throw off the test, since most of the length of the given time series would be (near-)stationary.
Similarly, if a corpus (equivalent to the `sampling period' in a genetics experiment) is too `short', it might only chronicle a segment of a longer change process.

\section{The FIT and binning decisions in linguistic corpora: a reanalysis of English past tense verb regularization}\label{section:reanalysis}

\begin{figure}[htb]
	\noindent
	\includegraphics[width=\columnwidth]{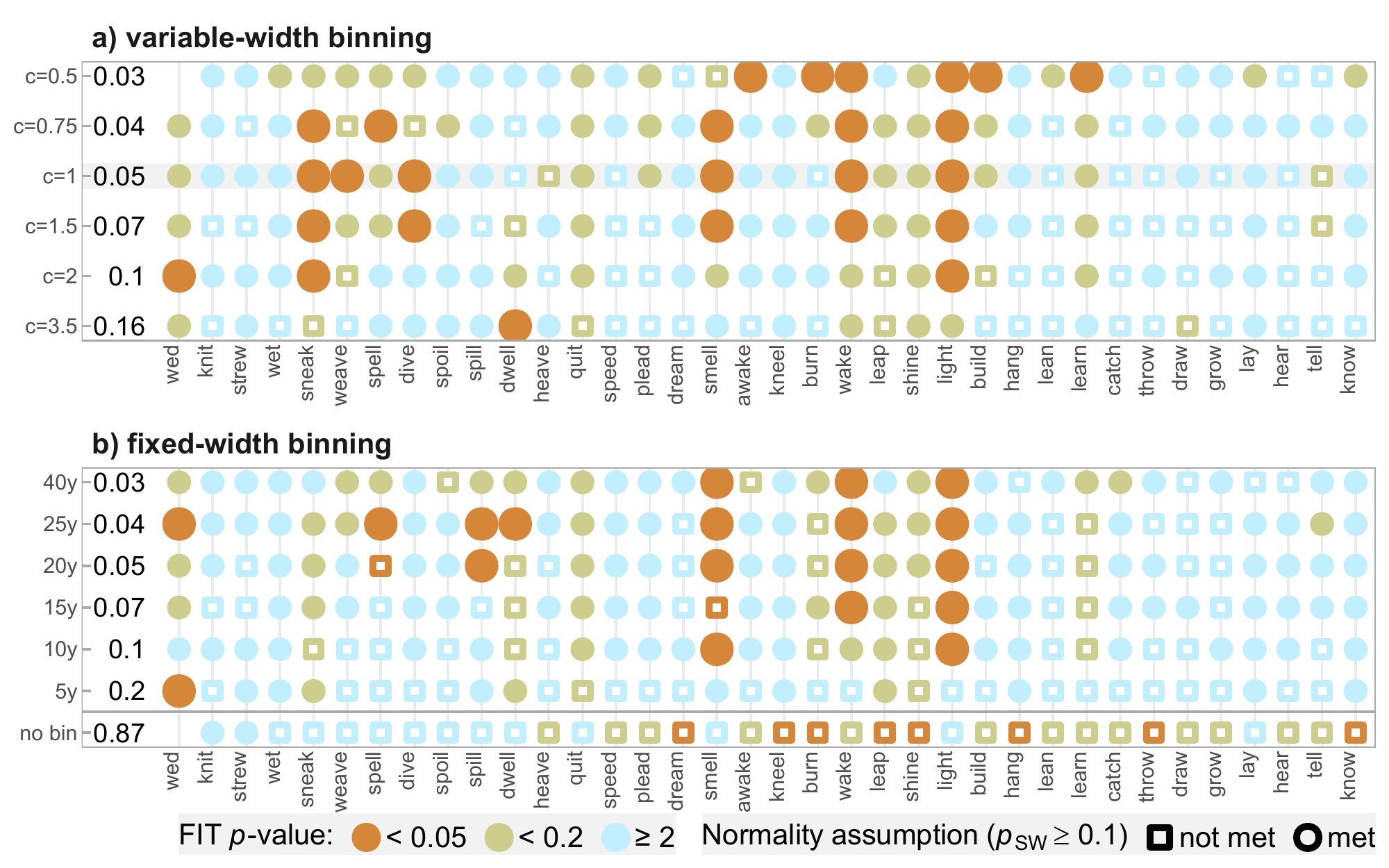}
	\caption{
		Results of applying the FIT to time series constructed based on 200 years of COHA frequency data. The verbs are ordered by overall frequency (low on the left). The constant $c$ determines the number of variable length bins via 
		$n(b) =  c\ln(n(v))$. $c=1$ corresponds to Newberry~et~al.'s original results.
		$10{\rm y}$ corresponds to fixed bin length of 10 years, etc; `no bin` refers to no additional binning on top of the default yearly bins in the corpus.
		The colour of each point corresponds to the result of the FIT test of a verb time series in each binning (orange: $p<0.05$, gold: $0.05\leq p<0.2$, light blue: $p\geq0.2$). The shape corresponds to the Shapiro-Wilk test result (filled circle: $p\geq0.1$, hollow square: $p<0.1$, likely not normal), with cases of selection meeting the normality assumption highlighted by a larger circle. 
		The column of numbers on the left displays the (rounded) median of the bins to years ratio in the given binning strategy. Only years where the verb occurs are counted (exclusion of sparse bins also leads the median in the no-binning version to be below 1). The listed variable (panel a) and fixed-width strategies (b) yield comparable binning ratios, e.g. the ``$c=1$'' version is comparable to 20-year fixed-width.
		In summary, the results presented here demonstrate that the FIT is sensitive to the strategy used for binning}\label{fig:fitresults}
\end{figure}

We focus here on the main result of Newberry~et al.\ --- the application of the FIT for assessing time series of verb form frequencies in order to determine if the observed patterns of change for 36 English verbs results from stochastic drift or selection.
Technical data processing details described in this section are based on the Supplementary Information of Newberry~et~al., their code, and M.~Newberry, p.c.

They construct a time series for each of 36 pre-selected verbs using 200 years of data in the Corpus of Historical American English (COHA), by counting how many times the regular past tense form occurs relative to the total number of instances of either the regular or irregular form.
The yearly verb count series are then binned (grouped) into a number of variable-width quantile bins $n(b)=\lceil\ln(n(v))\rceil$, where $n(v)$ is the sum of both (regular and irregular) past tense form tokens of the verb counted across the entire corpus. 
For example, \textit{light}.\textsc{past} occurs $n(v)=8869$ times in the corpus, resulting in $\lceil\ln(n(v))\rceil=10$ bins to group the years where the verb occurs. The first bin contains years 1810-1863 (and contains 897 tokens), the second 1864-1886 (890 tokens) and so on, up to the tenth (1994-2009, 884 tokens). Since the grouping is by years (years being the time resolution of the corpus), the bin size varies slightly in the exact number of tokens falling into each bin.
More frequent verbs thus get more bins (up to 13), whereas less frequent verbs get fewer bins (down to 6). 
For each verb in each bin, the relative frequency of its regular past tense form in $[0,1]$ is calculated.
Since the FIT assumes relative frequencies in $(0,1)$, Laplace $+1$ smoothing is applied to count values in bins where one of the variants has no occurrences at all in this section of the corpus.

As discussed above in the section on corpus binning, \textit{some} temporal segmentation process is necessary. The binning procedure applied by Newberry~et~al.\ is somewhat different from the more common strategy of using fixed length bins such as years or decades. The advantage of their approach is that there is guaranteed to be data in every bin (whereas a low frequency lexeme might be entirely absent in a fixed-width bin), the bins are roughly the same size in terms of tokens, and the resulting increments tend (although are not guaranteed) to be normally distributed with equal variance. 
These properties are beneficial for the FIT, more likely yielding normally distributed increments with less sampling noise \parencite{feder_identifying_2014}. 
It should be noted though that the resulting bins differ quite widely in their temporal granularity --- e.g. in the example above, the longest bin covers the earliest 53 years of the corpus, the shortest covers the most recent 15 years, and different verbs will use different time windows depending on their frequency in the corpus. Since the COHA is smaller on the early end (less tokens per year) and bigger on the more recent end, variable-width bins of the verb data are systematically longer in the early 1800s compared to the 20\textsuperscript{th} century ones (cf. the Supplementary Appendix for more discussion).

The series of relative frequencies based on the resulting bins are fed into the Frequency Increment Test to assess whether one may reject the null hypothesis of drift and assert that a given trajectory is therefore probably a product of selection. Newberry~et~al.\ set the FIT $\alpha=0.05$ but also report results for $\alpha=0.2$. They conduct the Shapiro-Wilk normality test on the transformed frequency increments, as the FIT assumes the increments to be normally distributed.

We replicate their original results, using their code, and furthermore explore the consequences of manipulating the size of the bins, in two ways. We present results for both binning strategies. That is, variable-width bins, $n(b)=c\ln(n(v))$, where $c$ is an additional arbitrary constant, and $c=1$ recovers the Newberry~et~al.\ procedure; and fixed-width bins, each set to a fixed duration in years.

Importantly, the fixed-width binning approach necessitates the introduction of an additional parameter: since some bins may end up with no or few occurrences of either form of a verb, we set a threshold of minimum 10 total occurrences for a relative value to be calculated in a bin; otherwise the bin is excluded before applying the FIT (hence also reducing the number of bins that make up the time series).
As the FIT assumes values in $(0,1)$, smoothing of boundary values is required. But if there is only a single occurrence of a verb in a bin (meaning the single present form would be at 100\%, the other at 0), then the $+1$ smoothing would force the relative value to be 50-50, which is undesirable. Similar distortions would happen with small frequency values, hence the threshold of 10.
See the Supplementary Appendix for more discussion on the differences between these approaches and how different minimal frequency thresholds affect the results. A more conservative threshold (such as 100) would yield more reliable bins (and less noisy time points), but given the size of COHA, most verbs don't have 100 occurrences per year (or some even in 5 years), which would preclude testing in shorter fixed bins.

Figure \ref{fig:fitresults} shows the results of these various analyses, in terms of how many verbs (out of the 36) allow us to reject the null hypothesis of drift, given the thresholds mentioned in the original work, as well as taking into account the normality assumption of the FIT (see above). We use the Shapiro-Wilk normality test, following Newberry~et~al.\ (this test is of course subject to low power in small samples as well). Out of the 466 time series analyses summarised in Figure~\ref{fig:fitresults} (36 verbs times 13 binning choices, minus two series with not enough data points), 63\% of the FIT $p$-values are eligible to be interpreted at Shapiro-Wilk $\alpha=0.1$.

We find that binning strategy does have an effect on the results, both in variable and fixed binning.
Importantly, in broad strokes, the picture presented by Newberry~et al.\ holds. They found that 6 out of 36 verbs undergoing selection; since the majority of verbs do not give a positive signal for selection, they interpret this as indicating that language change is often primarily stochastic. 
Looking at a wider range of binnings, we find that in most cases, there are indeed $5\pm2$ verbs that get flagged as undergoing selection at FIT $\alpha=0.05$, consistent with their conclusion. 
However, the specific verbs that are flagged as undergoing selection vary depending on the binning strategy. There are 4 verbs for which selection is detected in most binning choices --- \textit{light, smell, sneak, wake} (incidentally the ones with the strongest inferred selection coefficient, given the original binning, cf. EDT1 in Newberry~et~al.). 
There are also between 9 and 11 verbs (in variable-width binning; depending on how stringently the normality assumption is observed) which provide a robust absence of significant indications of selection, where the FIT $p$-value never drops below $0.2$ regardless of binning. However, for the remaining verbs the decision as to whether or not they are undergoing selection depends on the binning choices.
That being said, Newberry~et al.\ do draw attention to the fact that results of applying the FIT come with a certain margin of error and report their false discovery estimates (30\% for verbs with a FIT $\alpha=0.05$, 45\% at $0.2$).

Given that binning leads to different sample sizes of increments for the underlying $t$-test, those in turn being based on differing distributions of the tokens, some variance in the $p$-values is to be expected (not unlike in a replication of an experiment). The interpretation of our results and the appropriate conclusion regarding the sensitivity of the FIT test to binning strategy ultimately depends on one's intention in carrying out a tests of selection in the first place. If the goal is to test a large set of series to determine general tendencies, as is the case for Newberry~et al.\, then this approach may well be good enough --- the qualitative result of Newberry~et al.\ does broadly apply in most binning strategies.

However, most individual time series seem rather sensitive to binning, in the sense that the $p$-values fluctuate across conventional $\alpha$ levels between binnings. No verbs show an unambiguous signal of selection. 
For example, drift is not rejected in the time series of \textit{wed} using the Newberry~et~al.\ binning, while it is when the number of variable-width bins is multiplied by 2. 
The verb \textit{sneak} is significant at $\alpha=0.05$ in almost all the variable-width binnings, but in none of the fixed length ones; \textit{awake} is significant in only a single explored binning strategy (variable-width with $c=0.5$) and there are 4 more such verbs particularly sensitive to binning (the 1-year bins notwithstanding).

The no-binning results (i.e., using the default 1-year bins of COHA without further binning) differ visibly from the rest, but the normality assumption is also mostly violated. Given the small and variable bin sizes (tokens per bin), the same is likely true for the homoskedasticity assumption (although how much that matters and how to set a threshold is not clear, cf. the Supplementary Appendix). Most importantly, using `default' 1-year bins leads to testing on series where the increments are often based on very small samples, which is not desirable for any statistical test.

These evaluations obviously depend on the choice of $\alpha$ thresholds for the FIT and the supporting normality test --- for example, a more stringent FIT $\alpha$ would lead to more verbs being classified as unambiguous cases of drift.
In any case, if the intention is to test a particular example of linguistic change for selection (something a linguist may well be interested in), things become difficult. The issue diminishes if there is sufficient data on the variants, but that does not seem to be the case for many of the verbs tested here, given the size of COHA.

All in all, these findings merit a further investigation into the inner workings of the Frequency Increment Test and its applicability to corpus-based time series, which we will conduct in the following two sections.

\FloatBarrier

\section{The behaviour of the Frequency Increment Test in artificial time series}\label{section:smallsamples}

\begin{figure}[htb]
	\noindent
	\includegraphics[width=\columnwidth]{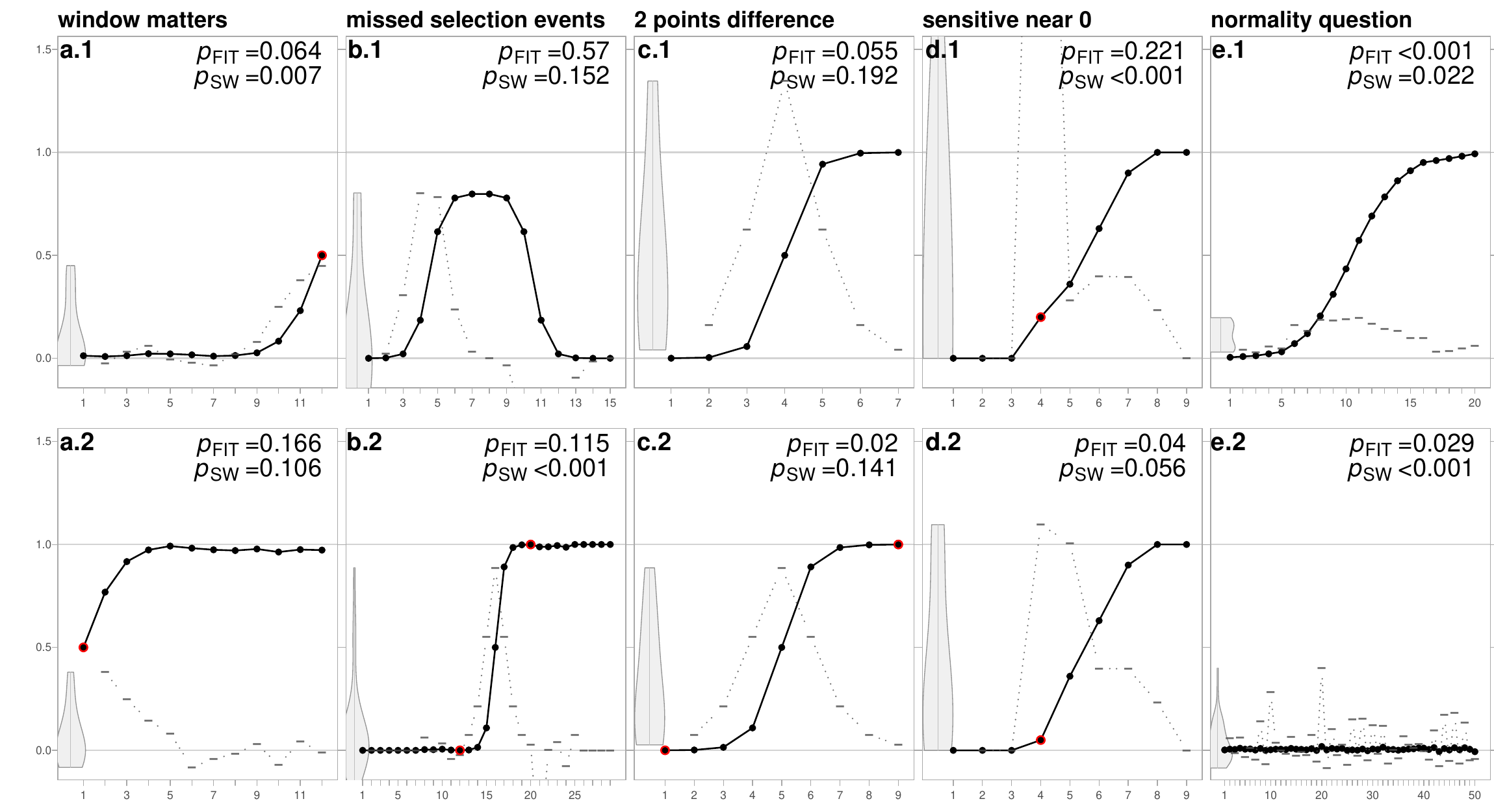}
	\caption{
		Artificially constructed time series of fictional variant relative frequencies (thick black lines, in $(0,1)$); time on the x-axis. The rescaled increments (after adjusting for absorption) are shown as dotted grey lines with dash points, and their distribution is shown on the left side as a violin plot. Points of interest discussed in this section are highlighted with red on some panels. The FIT and Shapiro-Wilk test $p$-values are reported in the corners.
		This figure depicts a number of realistic scenarios where applying the FIT would yield unexpected results, due to either the range of the time series derived from the corpus (a, b), a difference in the number of data points (c),  the sensitivity of FIT to near-zero values (d, e), and how stringently the assumption of the normality of the distribution of increments is being observed (e). This figure illustrates reasons to exercise caution when applying a test like the FIT to linguistic time series}\label{fig:fitexamples}
\end{figure}

We construct a number of artificial examples (Figure~\ref{fig:fitexamples}) to probe the behaviour of the FIT on time series of length and character similar to those investigated in the original paper (which contained between 6 and 13 time points).
The FIT can be shown to yield robust results for a certain range of series (as already shown by the subset of binning-insensitive verbs in the previous section). 
Yet we also observe a number of scenarios --- time series that could be plausibly derived from linguistic corpora --- where the results of the FIT are perhaps not what one might expect, from a language science point of view.
To put it another way, this is the section where we push the FIT and see if it breaks.
The next section demonstrates scenarios where the results of the FIT remain robust.

Each series in Figure~\ref{fig:fitexamples} may be interpreted as the percentage of a variant of some fictional linguistic element over time (after binning). We calculate the FIT $p$-value of each series, as well as the Shapiro-Wilk test $p$-values.
Figure \ref{fig:fitexamples}.a draws attention to how the temporal range of the time series (or that of the coverage of the corpus) can lead to quite different conclusions. 
Both \ref{fig:fitexamples}.a.1 and \ref{fig:fitexamples}.a.2 are different ends of the same series (the overlap highlighted with the red circle). The series, if analysed as a whole, would yield a $p_{FIT}=0.02$, but neither end on its own holds sufficient data to reject drift (nor is the FIT technically applicable, if the assumption of normality is observed).
This perspective may explain the case of the purportedly drift-driven regularization of the verbs \textit{spill} and \textit{burn}, which are brought up in Newberry~et al.\ as examples where drift alone is sufficient to explain the change, but which are problematic because the regular forms were already highly frequent by the early 19th century where the COHA coverage starts. \textit{spill} starts out with a share of 55\% regular forms in the first bin given the variable-width binning strategy; \textit{burn} is at 86\% regular. Under fixed decade binning, \textit{burn} is 36\% regular in the first bin, increasing to 62\% and then to 82\%, indicating a sharp increase characteristic of strong selection rather than drift (but obscured by the variable binning approach). 

This example also points to a case where different evolutionary domains (genetics, language) might have different expectations about what a reasonable time-series characteristic of selection should look like. The FIT assumes the Wright-Fisher as the underlying model (reasonably so in population genetics). The long tail of near-zero values followed by a sudden increase in \ref{fig:fitexamples}.a.1 is something that is unlikely to be observed in a Wright-Fisher model with constant selection strength parameter. 
However, from a linguistic point of view, this is a very natural series: a recent innovation or borrowing will be represented in the corpus as an increase preceded by a period of zero frequencies as far back as the corpus goes; this pattern could be explained as a recent change in fitness (e.g. a change in the subjective sociolinguistic prestige of a word).

A similar case is presented in Figure~\ref{fig:fitexamples}.b.1: if the time series chronicles both strong selection for one variant, and subsequent selection for the competing variant, then a blind application of the FIT will invariably indicate drift. Using only (either) half of the series as input to the test would yield a $p$-value indicating selection. \textit{knit} is a verb undergoing a somewhat similar process, with usage spiking towards the regular (observable under finer binnings), followed by mostly irregular usage. 
Figure \ref{fig:fitexamples}.b.2 is an example of the behaviour of FIT if the corpus coverage is \textit{too} wide. The S-curve in the middle would yield a FIT $p$-value of 0.02 --- in fact, it is the exact same curve as in Figure~\ref{fig:fitexamples}.c.2 (highlighted by the red dots). Yet the~S being surrounded by (near-)absorption values, the FIT would indicate drift (were the test to be used despite the possible non-normality of the distribution).

In the case of real data, the part of the time series depicting the long period of no change could in principle be clipped away. This is straightforward if the ``tail" consists of zeroes, but less so given small near-boundary values. Similarly, only the part of the time series far enough from the boundaries could be analysed (keeping in mind the specifics of the FIT, see above). However, any such solutions would introduce yet another researcher degree of freedom (what part of the series to include in the analysis) \parencite[cf.][]{simmons_falsepositive_2011}.

Figure \ref{fig:fitexamples}.c further illustrates how the FIT result is affected by a change in the way the time series is operationalised (e.g., using a different number of bins). \ref{fig:fitexamples}.c.1 and \ref{fig:fitexamples}.c.2 are S-curves with identical parameters, differing only in length (by 2 data points). Yet their FIT $p$-values are notably different (see the next section for more on sensitivity to binning differences). As expected, the FIT is sensitive to small changes if the sample is small (being based on the $t$-test). This may explain to some extent the changes in FIT $p$-values of short time series, between similar binnings differing only by a few points in length (cf. Figure~\ref{fig:fitresults}). However, fewer bins can also lead to a lower $p$, if it results in a less jagged time series (likely the case for e.g. \textit{burn}; cf. Section \ref{section:wfsimulations} for the effects of binning on drift series).

The examples so far however have had more to do with particularities of pre-test data manipulation.
Figure~\ref{fig:fitexamples}.d illustrates a property of the FIT, its sensitivity to changes near the boundaries. \ref{fig:fitexamples}.d.1 and \ref{fig:fitexamples}.d.1 differ only by the value of the fourth data point, but the resulting FIT $p$-value is quite different (and furthermore the Shapiro-Wilk test indicates departure from normality in the increment distribution due to the outlier). 
The issue of applicability of the FIT to series with increments departing from normality is further illustrated with the last pair of series. \ref{fig:fitexamples}.e.1 is a typical S-curve often observed in language change, but the non-normal distribution of its increments would disallow the interpretation of the FIT $p$-value (that would otherwise indicate a clear case of selection). 

We observe that in general, for longer series exhibiting monotonic increase (characteristic of strong selection), the distribution of the increments quickly veers into the non-normal (as indicated by the Shapiro-Wilk $p$-value; other normality tests behave similarly; see also the Supplementary appendix). 
Time series composed of random values drawn from a uniform or normal distribution (or log-normal with  small $\sigma$) --- i.e., the kind of series that should exhibit no selection --- tend to have increments distributed approximately normally, as long as the series is away from the boundary values. However, the increments of S-shaped curves tend towards a bimodal distribution. Increment distributions of are severely skewed when a series is shaped like an S-curve but with a sharp ``bend", a straight line (linear increase or decrease), and when a series include long periods of no change.

The assumption of normality could of course be relaxed. However, we observe that this would lead to at least one additional issue, in the form of false positives stemming from the sensitivity of the FIT to small near-boundary changes, illustrated by \ref{fig:fitexamples}.e.2. Given a long enough series of random values (here sampled from a normal distribution) with a near-zero mean and small standard deviation, the FIT often yields a small $p$-value 
(the same applies to samples from the uniform and log-normal distributions; this effect is not observed when the mean is away from the boundaries). Such series would however usefully get flagged as having non-normal increment distributions.

This is also likely why the otherwise flat-lining series for \textit{tell} in Newberry~et~al.\ ends up being included in the discussion as a possible case of selection (at FIT $p=0.12$, with a red flag of Shapiro-Wilk $p=0.001$). Among the 12 bins of its series (under the original variable-width quantile binning procedure), it has only a few once-per-bin occurrences of regular \textit{telled} after the initial three bins --- a total of 4 singleton occurrences spread out over the span of a century. 
The $+1$ absorption adjustment forces the zeroes for \textit{telled} in the rest of the bins to be ones as well. The observed fluctuations (and resulting FIT $p$-value) in the series only reflect the slightly fluctuating token frequency of \textit{tell}, which ranges between 9189 and 11940 in the variable-width bins. Keeping the relative frequency value constant after the third bin instead (at the value equal to the third bin to avoid bias) would result in a FIT $p=0.21$.

These last four usages of the regular past form \textit{telled} in COHA all occur in the fiction part of the corpus, all appearing to reflect the intention of the author to convey a particular kind of character (not used randomly as per a drift model).
This would be an example of how an archaic variant can re-surface --- quite possible in a language with a long written record, where speakers need not necessarily even directly ``inherit'' a variant from the previous generation.
In that case, \textit{telled} could be said to have been selected for, due to having increased fitness in a specific (stylistic) niche, and its usage is not due to random variation in the utterances of the speakers (or drift). However, as shown above, this possible (occasional) selection is not what the FIT is picking up on in this case, but rather simply the fluctuating frequency of \textit{tell}.

Meaning change can also give rise to apparent re-emergence of variants. The occurrence of a form does not guarantee that it is being used in the same meaning or function that it had in another period or context 
(an implicit assumption in Newberry~et al.). 
For example, the aforementioned \textit{spill} in COHA quickly converges to the regular past tense \textit{spilled}, but occasional usages of the irregular \textit{spilt} still occur, yielding what appears to be a randomly fluctuating time series. On closer inspection, the latter appear to be mostly adjectival usages, not actual past tense verbs, and often turn up in the lexicalized (or `fossilized') phrase of \textit{cry over spilt milk}.
Examples like that of the time series of \textit{telled} and \textit{spilt}, or the series in Figure \ref{fig:fitexamples}.a.2 and e.2. may possibly be seen as edge cases from the perspective of population genetics --- the original domain of the Frequency Increment Test and related approaches. However, as highlighted here, they are examples of not particularly uncommon processes (lexicalization, stylistic usage of unusual variants) in the domain of language.

Finally, one might argue the examples in Figure~\ref{fig:fitexamples} are not really counterexamples to the utility of the FIT, being representative of cases where the FIT is, strictly speaking, not designed to apply in the first place, such as series with not-quite-normal increments, long flat segments, and values near the boundaries. 
Excluding these however would mean excluding a fair share of language change scenarios easily observable in corpora, such as changes starting at zero as in cases of linguistic innovations, ongoing changes stretching beyond the bounds of a corpus, and many S-curves typical of language change (and series in general where the underlying selection coefficient is likely not constant). 
Yet dismissing these as invalid points of concern would also mean dismissing the FIT as a broadly applicable test of selection for the domain of language change. 

In the next section, we turn to simulations to explore the behaviour of the FIT beyond that of a few specific series (Section \ref{section:wfsimulations}), before finally trying to reconcile these conflicting viewpoints (Section \ref{section:discussion}).

\section{The effect of binning frequency data for time series: a simulated example}\label{section:wfsimulations}

Here we attempt to further explore the ``parameter space'' of applying the FIT to simulated data with known properties of selection strength and binning. (code to replicate these results: see the Data availability section in the end).
We use the Wright-Fisher model \parencite{ewens_mathematical_2004} to simulate a large number of time series using the following parameters: population size $N=1000$ ($N$ here does not refer to the ``population" of speakers, but is analogous to the sum of parallel variants in a corpus bin, e.g. the sum of the counts of \textit{lit} and \textit{lighted} in a given year); selection coefficients $s \in [0,5]$; 200 generations (the latter emulating COHA, where the minimal time resolution is 1 year, and there is 200 years of data). 
The update rule for this model is as follows. Given $n_t$ ``mutants'' (e.g., regular past tense forms) in generation $t$, each individual in the next generation is a mutant individual with probability 
\begin{equation}
q = \frac{n_t(1+s)}{n_t(1+s) + (N-n_t)}
\end{equation}
Otherwise, it is the wild type (e.g., irregular past tense forms). Where $s=0$ we have random drift; higher values of $s$ given an increasingly strong selective advantage to the mutant variant.

Each series (200 data points) is binned into a decreasing number of bins (i.e., $[200, 4]$, of length $[1,50]$), and the FIT is applied to every binned version. The simulation for each combination of selection strength and bin length is replicated 1000 times. 
In summary, in this section we vary the selection strength $s$ and binning, while keeping $N$ and the number of generations constant.

Importantly, we also apply binning to the series post-simulation the same way one would apply binning to corpus counts, as discussed above.
The obvious difference from corpus-based time series is that the latter usually do not come from a population with a stable size (total lexeme frequency usually varies in addition to variation in its variants), and are often not continuous (gaps where a lexeme might be completely absent). Since our artificial series do not suffer from these problems, variable-width and fixed-length binning yield identical results, and we can simply use the latter.

We explore two scenarios, where the competing ``mutant'' variant starts out at 50\% of the population and where it starts out at 5\%. The former is useful for exploring the effects of binning at low $s$ and false positive rates, the latter for exploring high $s$ and false negatives.
Obviously, any specific $s$ thresholds and ranges discussed in this section apply to this specific experiment and would likely be somewhat different given series of different length and $N$ (cf. the Supplementary Appendix for some further exploration).

\subsection{Drift and low selection}

\begin{figure}[htb]
	\noindent
	\includegraphics[width=\columnwidth]{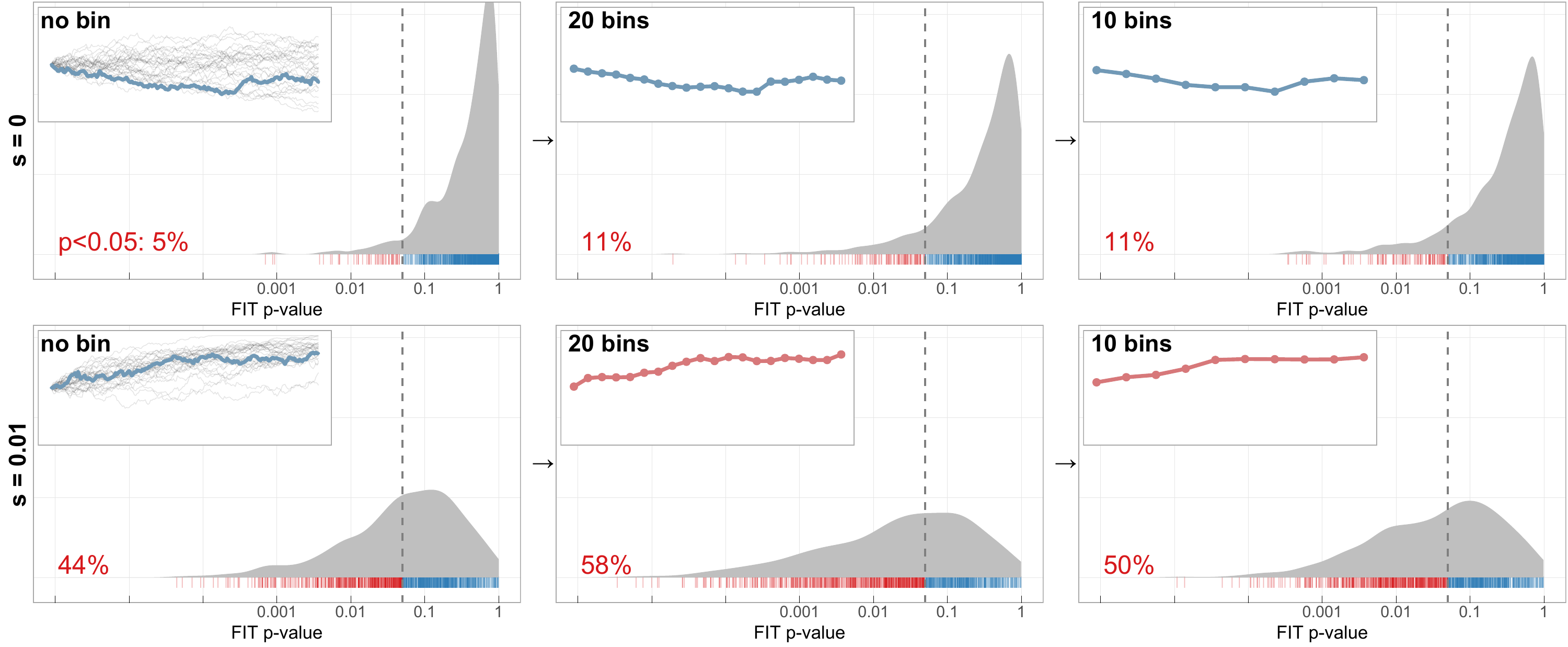}
	\caption{
		The distribution of FIT $p$-values given 1000 series from the Wright-Fisher model (200 generations, starting at 50\%). The panels are arranged from left to right reflecting increased binning. The small inset panels display how binning affects a single example series.
		$p$-values below 0.05 are coloured red (left of the dashed line), above 0.05 in blue. Note the $\log_{10}$ x-axis.
		This figure illustrates that the false positive rate is susceptible to increasing when the series are binned (top row). 
		At non-zero but low $s$, differences between binning and no binning can be more pronounced (bottom row).
		See Figure \ref{fig:fitmap2} for the full exploration of the parameter space}\label{fig:fitmap1}
\end{figure}

\begin{figure}[htb]
	\noindent
	\includegraphics[width=\columnwidth]{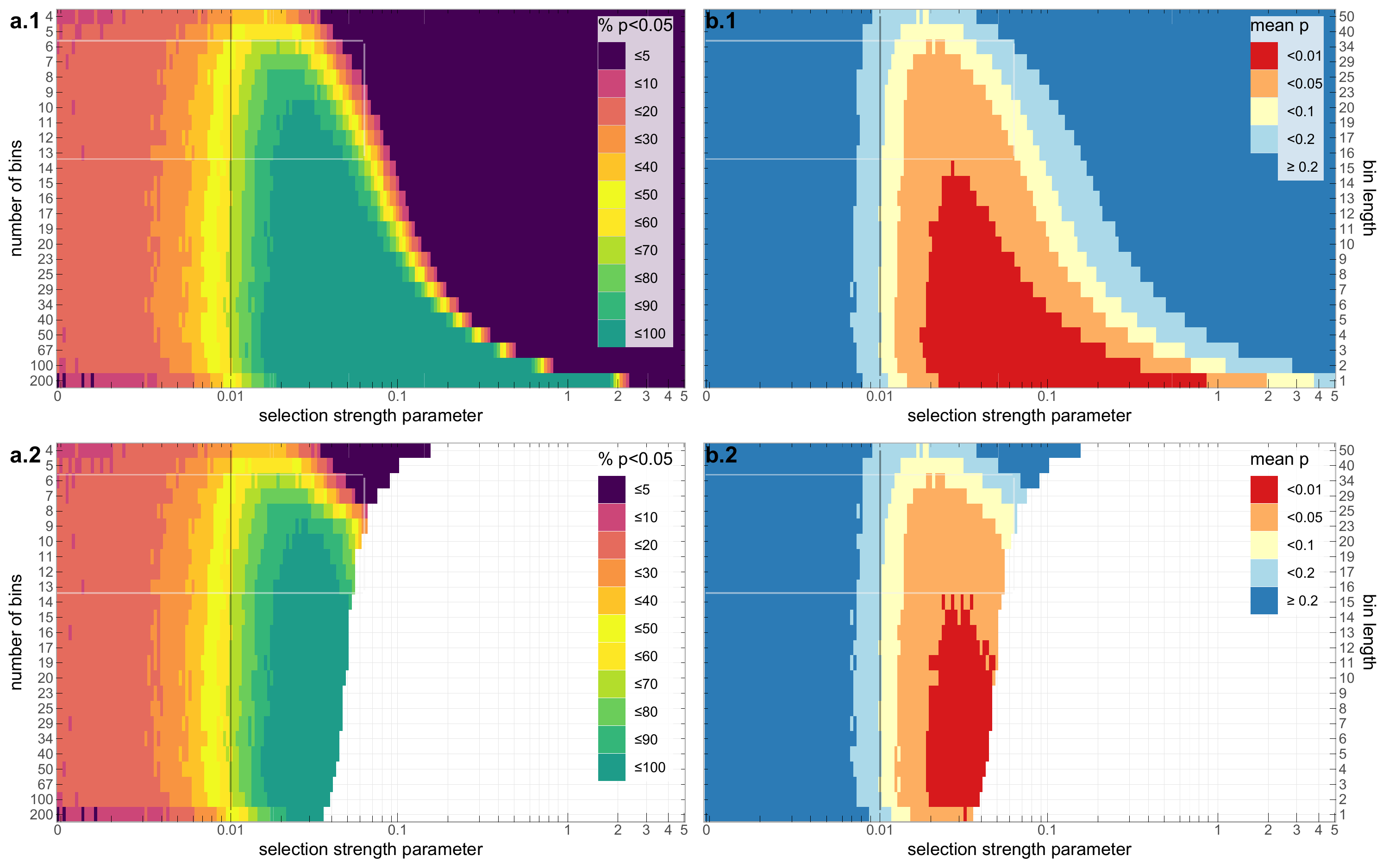}
	\caption{
		FIT $p$-values of time series generated using the Wright-Fisher model (with the ``mutant'' variant starting at 50\%), across a range of selection coefficients (x-axis, note the log scale), binned into a decreasing number of bins (y-axis). 
		Left in pink and green (a): \% of time series with FIT $p<0.05$, in 1000 replicates. 
		Right in red and blue (b): mean FIT $p$-value.
		The bottom pair (a.2, b.2): the same data, but series with a Shapiro-Wilk $p<0.1$ have been removed before calculating the percentages and means.
		The white rectangle: the range of $s$ and binning explored in Newberry~et al.
		The vertical black line highlights the $s$ explored in Figure~\ref{fig:fitmap1}.
		A consistent colour across a column of pixels indicates robustness to binning choices under the corresponding $s$, while variable colouring indicates sensitivity to binning}\label{fig:fitmap2}
\end{figure}

Figure \ref{fig:fitmap1} depicts how the results of the FIT change depending on binning, given a time series with low selection ($s=0.01$, bottom row) and no selection ($s=0$, top row; corresponds to the leftmost column of pixels on the panels in Figure~\ref{fig:fitmap2}).
At zero selection, the FIT has a reasonable false positive rate of around 5\% at $\alpha=0.05$. 
Binning such series into a smaller number of bins causes an increase in the share of $p$-values below 0.05 (presumably because noise is smoothed out). Binning appears to affect the $s=0.01$ range even more (bottom row).

Figure \ref{fig:fitmap2} represents the entire parameter space explored in this experiment for the 50\% start condition. Each pixel on the heat maps corresponds to a parameter combination of selection strength (horizontal axis) and number of bins (vertical axis). 
The vertical axis starts with 200 or no binning, corresponding to bin length 1 --- and running up to 4 bins, with bin length 50, being the result of 200 data points squeezed into the 4 bins. 
Minimal binning --- compressing 200 generations into 100 bins of length 2 --- appears to make the clearest immediate difference: the share of $p<0.05$ is consistently about 10\% higher between the binned and non-binned series when $s$ is low (observe the bottom two ``shifted" looking pixel rows in Figure~\ref{fig:fitmap2}.a.2).

The 50\% start is suitable for exploring low selection, as in the case of lower starting values, many such series hit absorption or ``run into the ground", and the resulting mostly-zero series would violate the normality assumption (of its underlying Gaussian approximation of the diffusion process).
However, the higher $s$ range in Figure~\ref{fig:fitmap2}.a.2 could be interpreted as a model of the situation where a change is only partially chronicled by a corpus, e.g. Figure~\ref{fig:fitexamples}.a.2 in Section \ref{section:smallsamples}.
Selection becomes understandably difficult to detect in very short series regardless of the underlying selection coefficient.

\FloatBarrier

\subsection{High selection}

\begin{figure}[htb]
	\noindent 
	\includegraphics[width=\columnwidth]{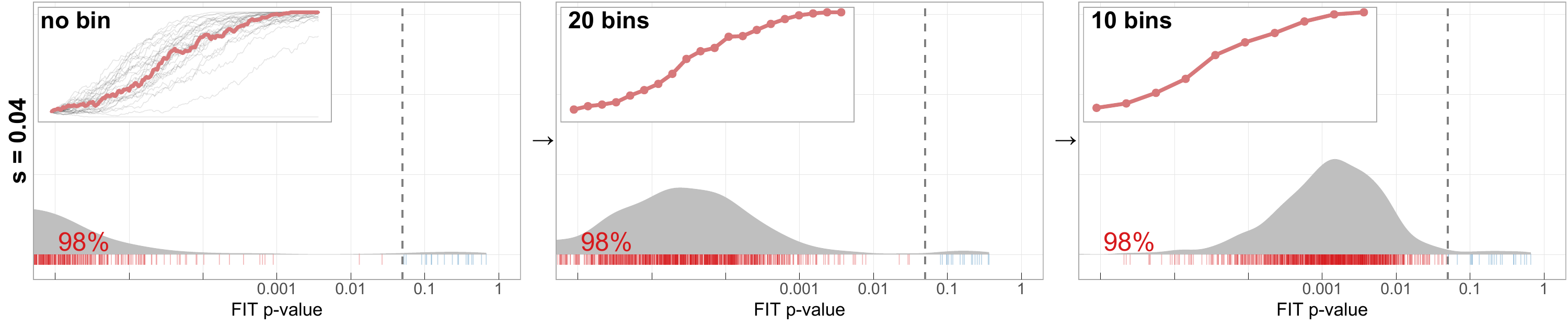}
	\caption{
		The distribution of FIT $p$-values given 1000 Wright-Fisher series with strong selection (200 generations, starting at 5\%). The panels are arranged from left to right reflecting increased binning. The small inset panels display how binning affects a single example series. 
		$p$-values below 0.05 are coloured red (left of the dashed line), above 0.05 in blue. Note the $\log_{10}$ x-axis.
		The red value in the bottom left corner shows the percentage of $p$-values below 0.05.
		This figure illustrates the $s$ range where the FIT is most robust to binning, retaining a small and stable false negative rate (i.e. the inverse of the percentage value in the corner)}\label{fig:fitmap3}
\end{figure}

The 5\% start is suitable for exploring high selection, as with higher starting values, many high-selection time series reach absorption fast, yielding series not meeting the increment normality assumption.
Figure \ref{fig:fitmap3} depicts distributions of FIT $p$-values under different binnings, given time series with a moderately high $s$ of $0.04$, and the incoming variant starting out at 5\%. 
This appears to be the subset of series where the FIT works very well and is most insensitive to binning choices.

\begin{figure}[htb]
	\noindent
	\includegraphics[width=\columnwidth]{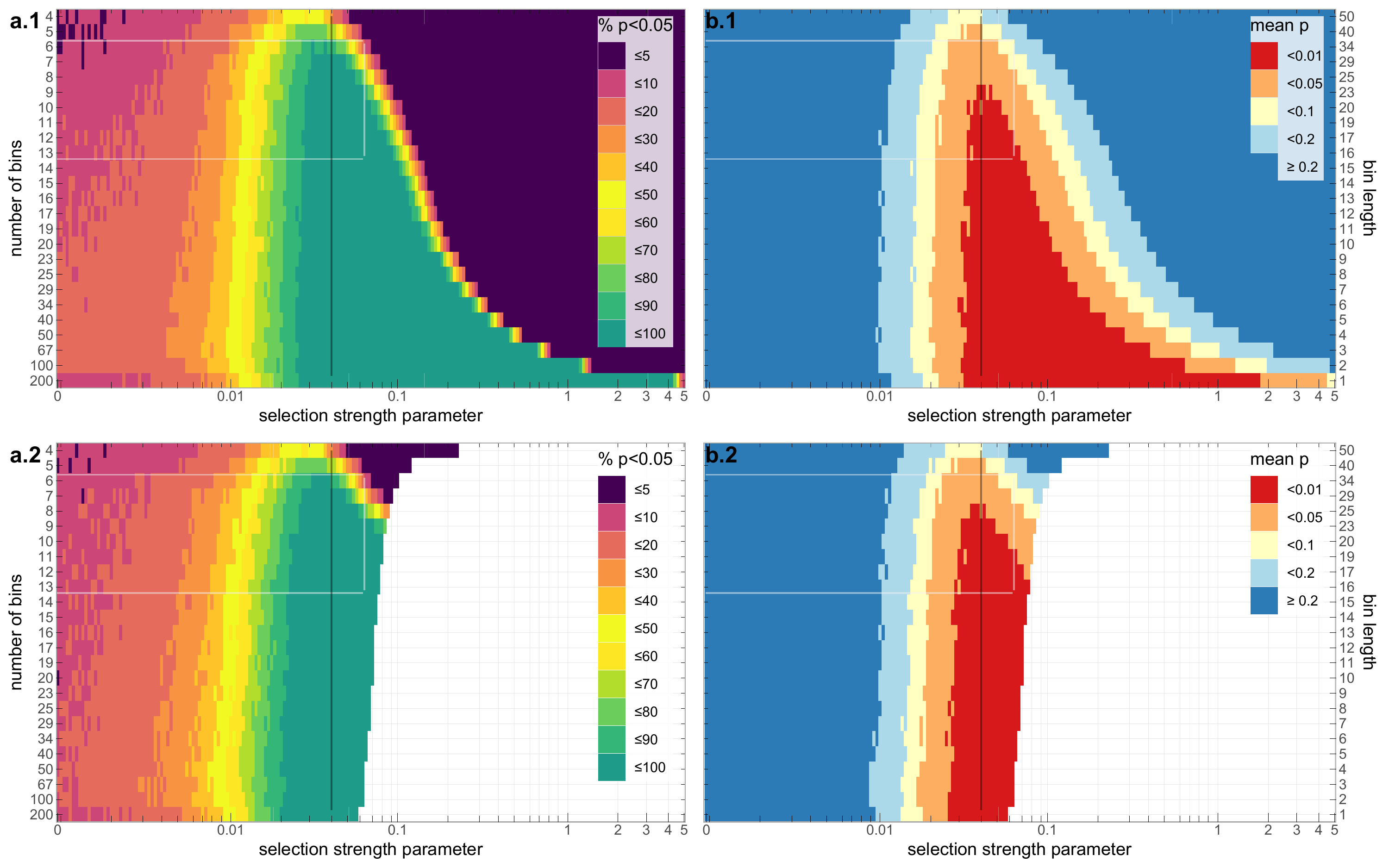}
	\caption{
		FIT $p$-values of time series generated using the Wright-Fisher model (with the ``mutant'' variant starting at 5\%), across a range of selection coefficients (x-axis, note the log scale), binned into a decreasing number of bins (y-axis). 
		Left in pink and green (a): \% of time series with FIT $p<0.05$, in 1000 replicates. Right in red and blue (b): mean FIT $p$-value.
		The bottom pair (a.2, b.2): the same data, but series with a Shapiro-Wilk $p<0.1$ have been removed before calculating the percentages and means.
		The white rectangle: the range of $s$ and binning explored in Newberry~et al.
		The vertical black line highlights the $s$ explored in Figure~\ref{fig:fitmap3}.
		A consistent colour across a column of pixels indicates robustness to binning choices under the corresponding $s$, while variable colouring indicates sensitivity to binning}\label{fig:fitmap4}
\end{figure}

Beyond that, things become more complicated. Our reanalysis of the 36 verb time series in Section \ref{section:reanalysis} indicated that it is series exhibiting the strongest selection that would remain consistent in terms of their FIT result across the different binnings. However, as illustrated in Figure \ref{fig:fitmap4}, it seems too high selection can have the inverse effect, as this is where false negatives begin to crop up under too much binning (e.g. with 10 bins, $>10\%$ at $s=0.07$, $>90\%$ at $s=0.1$). 
That is, if the increment normality assumption is being be strictly observed --- if it is, then the results of the test are not valid any more at this range (cf. white area in Figure~\ref{fig:fitmap4}.a.2).
This illustrates that the FIT has a maximum selection strength for which it is effective. 
At higher selection strengths, i.e. above $0.06..0.1$ in our toy model, sensitivity to binning and violations of the normality assumption both become problematic, yielding results with a high false negative rate (if the assumption is relaxed; cf. Figure~\ref{fig:fitmap4}.a.1) or results which are invalid (if it is observed; \ref{fig:fitmap4}.a.2). Incidentally, this also is the $s$ range where S-curves characteristic of language change begin to form (cf. the Supplementary appendix).

In summary, these results indicate that if one is to take the same ensemble of language changes, with known selection strength, and apply different binning protocols, one could easily end up drawing very different conclusions depending on the bin length and the normality assumption threshold, if the conclusions are based solely on applying a test such as the FIT.
However, if awareness of these limits is maintained, then the FIT works well on time series with moderately strong selection, and reasonably well (with the caveat of somewhat increased false positives rate under binning) on time series generated by a zero or low selection coefficient.

\FloatBarrier

\section{Discussion}\label{section:discussion}

We started out by focussing on the study of the (ir)regularisation of the past tense of 36 English verbs in Newberry~et~al.\, specifically their finding that drift cannot be rejected in most cases, leading to the claim of the ``an underappreciated role for stochasticity in language evolution'' \parencite[][223]{newberry_detecting_2017}.
The conclusion of our reanalysis section --- that their broad conclusion stands but that the FIT is sensitive in specific instances to the chosen binning strategy --- prompted further investigation of the properties and range of potential applicability of the FIT.
In the following sections, we demonstrated that the FIT yields reasonable results in a certain subset of possible time series, yet perhaps less expected results in others, when applied to a variety of series with different lengths, shapes and underlying selection coefficients.

The fundamental issue is that corpus data has to be operationalised one way or another if one is to apply a time series analysis that is based on variant frequencies. There is as yet no single best method to do so, and the additional researcher degree of freedom is practically unavoidable. 
Also, unlike microbial experimental data --- for which the FIT was designed originally --- the beginning and end of a corpus in terms of temporal coverage may not necessarily overlap with the beginning and end of a language change trajectory. The implications of these scenarios on the FIT approach were explored in Figures~\ref{fig:fitexamples}~and~\ref{fig:fitmap2}. 
Any test based on increment signatures is likely to miss a significant change, if it is recorded by very few data points. This could be either due to data sparsity or low number of bins, very high underlying selection, or the change happening in the middle of an otherwise long series. 
This could be remedied to an extent by only considering the bins of a corpus or the segments a time series where a change ``looks like'' it is taking place --- but that introduces yet another parameter or researcher degree of freedom.

In what follows, we attempt to summarize our findings and distil them into actionable guidelines for applying tests of selection to linguistic corpus-derived time series.

\subsection{Limitations for linguistic selection testing}

Besides the fact that caution should be exercised when its statistical assumptions are not met (as with any statistical test), the following should be taken into account when applying the FIT or a similar test of selection to corpus data. $s$ continues to refer to the selection coefficient driving the process of change (assuming an underlying Wright-Fisher like process; see Section~\ref{section:fittest} for related discussion).
Obviously, a test of selection being carried out implies that $s$ is actually unknown to the tester --- the guidelines sketched here are meant to draw attention to situations where it might be beneficial to inspect the results more carefully.
In terms of the input data quality, the results of a test can be misleading if the time series:
\begin{itemize}
	\item chronicle only a part of a change (beginning or end);
	\item are too short (too few data points or bins);
	\item are too long (if covering multiple events, variable $s$);
	\item based on greatly variable bin sizes (avoidable with variable-width binning, which leads to variable bin lengths).
\end{itemize}	
In terms of the types and shapes of possible series, binning can lead to unpredictable results in the case of FIT (and its assumption of increment normality is likely violated) in time series:
\begin{itemize}	
	\item which are S-curves (non-normal increments);
	\item where $s$ may be suspected to vary over time (e.g. S-curves with long tails);
	\item where $s=0$ (binning increases false positives);
	\item with a very high $s$ (sharp changes, quick fixation);
	\item with tiny near-boundary fluctuations;
	\item where such values are introduced by smoothing (absorption adjustment).	
\end{itemize}
The high $s$ and absorption issue can be avoided by either excluding any series with a long span of zeroes or by making a choice to clip the post-absorption part of the series. That may leave a variable number of very few data points, and of course requires some consistent method of choosing the clipping point.
The tiny fluctuations issue is typically caused by occasional occurrences of the less popular variant of a pair or set with a very high underlying total token frequency. Such series can be avoided by checking for the normality of increments.

As exemplified in this contribution, the way data is handled can in some cases drive the results of a test of selection. An application of such a test --- particularly if it is borrowed from a different domain --- should thus take into account the nature of the data. 
In the case of time series derived from diachronic corpora, a number of issues require attention. 
These include corpus size and normalisation \parencite{gries_useful_2010}, 
quality of corpus tagging (cf. Supplementary appendix),
genre \parencite{szmrecsanyi_about_2016} and topic \parencite{karjus_quantifying_2018} dynamics,
representativeness and composition \parencite{pechenick_characterizing_2015,koplenig_impact_2017,lijffijt_ceecing_2012}.
For example, imbalances in genre or register can easily lead to a drifty-looking series, if the usage of a variant differs between them. 
It is also not clear how the interplay of multiple, possibly opposing sources of selection (inherent properties of the variant, sociolinguistic prestige, top-down language planning, etc.) could be captured by a single test.
Properties inherent to language can make a difference, such as the aforementioned re-use of archaic variants from the written record (Section \ref{section:smallsamples}), or meaning change, which may reasonably resolve competition between variants as they go on to inhabit different niches \parencite[automatic methods exist to detect the latter, cf. ][]{dubossarsky_timeout_2019}. 
This relates to the issue of determining what variants do and which do not actually compete with one other for the same meaning or function, often referred to in sociolinguistics as the problem of the envelope of variation \parencite[cf.][]{walker_variation_2010}.

\subsection{Opportunities for linguistic selection testing}

On the bright side, despite these concerns, the Fitness Increment Test and presumably similar tests are likely reliably applicable to time series derived from linguistic corpus data when:
\begin{itemize}
	\item the series covers the entire change (yet if possible also excludes near-boundary values);
	\item the assumptions of the test are checked for;
	\item the underlying $s$ can be assumed to be constant;
	\item the interplay of $s$ ranges and binning is taken into account (simulations help);
	\item the corpus is large, representative and consistently balanced over time for genre, style and topics;
	\item the target token count for each time bin is large ($\gtrsim 100$, cf. the Appendix);
	\item the semantics of the pair (or set) of variants remain the same;
	\item the set of variants yielding the relative frequencies can be assumed to be competing, and the set contains all the competitors for a meaning or function.
\end{itemize}
Besides these rules of thumb, it would be beneficial in most cases to have some principled mechanisms to:
\begin{itemize}
	\item evaluate multiple possible binning choices for the robustness of the test results;
	\item deal with the ``leftover" flat part of the series before and after the change being analysed;
	\item distinguish drift and the effects of variable $s$ over time.
\end{itemize}
\smallskip

Possible use cases in linguistics involving the FIT (or a similar test) presumably fall on a spectrum where on the one end the subject of a study would be a single change in the history of a language, and the aim would be to determine if that change has occurred due to drift or due to individuals consistently selecting for one of the variants, owing to its perceived higher fitness. 
On the other end would be the evaluation of a very large set of linguistic time series derived from a corpus, with the aim to reveal general patterns and dynamics of language change processes. The study of 36 English verbs by Newberry~et~al.\ falls closer towards this end of the spectrum.

When the subject of a study is a single change (or a few), and the result hinges on a single test result, then we would naturally advise to take the preceding concerns into careful consideration, from data sampling and preparation to the specifics of a given selection test, while being mindful of the involved researcher degrees of freedom.
If a study veers toward the other end of the spectrum, involving a large set of series, then its design would largely come down to a choice between two approaches.

One could either take a ``big-data" approach, feeding the test with a very large set of time series to explore the role of selection and drift in language change, checking for only the minimal statistical assumptions of the test. The upside is that, hopefully, despite the concerns specific to corpora and language, true patterns would emerge, given enough data. The downside is of course the danger of garbage in, garbage out.

Or alternatively, one could take the approach of also trying to check for the various linguistic assumptions in addition to the statistical ones, filtering out unsuitable series. This would hopefully lead to better language science. On the downside, this requires the meticulous introduction of a number of extra parameters, or researcher degrees of freedom. Furthermore, the results might not be representative of general language change dynamics in the end, if based on testing only a niche subset of series ``suitable" for a given test --- of which there might not be that many either.
In other words, no free lunch.

\subsection{Future prospects}

The multitude of points listed above might sound like a lot of limitations.
However, we would not by any means conclude that efforts to detect selection in linguistic data should be abandoned.
The idea of detecting selection in diachronic linguistic data based on shapes or signatures is not new and remains an open challenge \parencite{sindi_culturomics_2016,reali_words_2010,blythe_neutral_2012,bentley_random_2008,amato_dynamics_2018}.
At the same time, methods for detecting selection continue being improved in the field of population genetics  \parencite{iranmehr_clear_2017,nishino_detecting_2013,terhorst_multi-locus_2015,schraiber_bayesian_2016}.

Perhaps it would be useful to draw a distinction between exploratory and confirmatory findings. In essence, this strand of research (including Newberry~et~al.) has remained exploratory.
Simulations with controlled properties allow for an evaluation of the performance of a test or model under various conditions and suspected confounds \parencite[cf. also][]{kauhanen_neutral_2017}.
However, to the best of our knowledge, there is currently no objective way
to evaluate such methods or compare their accuracy against one another,
in terms how well they reflect the actual selection biases operating on the level of the speaker, that may eventually give rise to a change in the consensus on the population level --- a sample of which is (the only thing that is) eventually observable in a diachronic corpus.
It would therefore be useful to distinguish between approaches that \textit{test} for selection, and those that more accurately generate (albeit potentially interesting and worthwhile) hypotheses. The latter may be useful e.g. when positing causes of language change --- be they linguistic, social, or cognitive in nature. If drift cannot be rejected, then theorising about possible ``causes'' of the change is unnecessary.

The difficulties with binning suggest that trying to manipulate the data to make it look more like the underlying Wright-Fisher model --- i.e., coarse-graining individual instances of use to construct the continuously-varying variant frequencies that the model predicts --- is not the way to go. An alternative procedure would be to include the process of sampling these instances of use to build the corpus as part of the model. For example, given some time series $x(t)$ generated by the Wright-Fisher model, then at an instant $t$ this model says that we should expect to encounter one of the two word variants with probability $x(t)$. In an ideal world, one would then maximise the likelihood of the observed sequence of tokens with respect to the parameters of the Wright-Fisher model (i.e., the selection strength and effective population size). This procedure looks to be somewhat computationally demanding, and may prove intractable for large corpora. However, such a procedure could in principle be applied to token counts as they appear in a corpus, without the need for pre-processing (such as binning) and the researcher freedom associated with it.

Another domain besides language which has attracted similar genetics-inspired modelling approaches is that of archaeology, particularly datasets of (pre-)historical artefacts \parencite{bentley_cultural_2003}. Similar concerns have followed: `time-averaged assemblages' of variants in cumulative cultural evolution (essentially binned data) can easily introduce bias in various tests \parencite{crema_revealing_2016,premo_cultural_2014}. 
Diachronic datasets (e.g. those based on the archaeological record, but similarly, corpora) only provide sparse, aggregated frequency information, which may be the reflection of a variety of neutral or selective transmission processes at the individual level \parencite{premo_cultural_2014,crema_revealing_2016,kandler_inferring_2017,kandler_analysing_2019}. Since these underlying processes cannot be directly observed (particularly in prehistoric data), \textcite{kandler_inferring_2017} suggest shifting the focus from identifying the single individual-level process that likely produced the observed data --- to excluding those that likely did \emph{not}. A corpus being a sample of individual utterances, this suggestion is worth consideration. Although the written record tends to have more metadata than the archaeological, the author of an utterance, along with their selective biases, is often unknown.

Detecting signatures of selection and drift in the evolution of language (and other domains of cumulative culture) remains an interesting prospect. It would be informative to see a comparison of the FIT-like selection detection methods that have been developed in population genetics or archaeology, applied to linguistic data, and systematically evaluated.
If the issues listed in the sections above could be solved, then this would certainly improve possibilities for exciting linguistic inquiry, inviting answers to questions such as, 
do lexemes experience stronger drift than syntactic constructions? 
What is the relationship of selection and niche \parencite{laland_cultural_2001,altmann_niche_2011} in language change?
Are some parts of speech more susceptible to change via selection than others? (M.~Newberry, p.c.)
What is the role of drift in creole evolution? \parencite{strimling_modeling_2015}
In semantic change? \parencite{hamilton_cultural_2016}
Are some languages changing more due to drift than others? \parencite[and if that relates to community size;][]{reali_simpler_2018,atkinson_speaker_2015}.
Can different types of selection be distinguished, e.g. top-down planning, grassroots \parencite{amato_dynamics_2018}, momentum-driven \parencite{stadler_momentum_2016}?

\section{Conclusions}

We find ourselves witnessing an exciting time for linguistic research, where more and more data on actual language usage is becoming available, encompassing different languages, dialects, registers, modalities, but also centuries. At the same time computational means for analysing big data have become readily accessible, hand in hand with the development of methods providing new insight into how languages function, change and evolve over time. Alongside and perhaps interlinked with these developments, language as a domain of scientific investigation has attracted interest in recent decades from fields traditionally not engaged in linguistic research, such as physics and biology.

We evaluated the proposal of \textcite{newberry_detecting_2017}, consisting of the application of the Frequency Increment Test as a method for determining whether any time series constructed from corpus frequencies of competing variants is a case of selection or a case of change stemming from stochastic drift. 
We found that while some of the original results remain robust to binning choices, other do not.
Based on constructed and simulated examples, we find that while the results of the FIT can be robust given a subset of suitable series, there are scenarios where they affected by the way the diachronic corpus data are binned.

We advocate that in the interest of reproducibility, binning, like any other data manipulation and operationalisation procedures, should be explicitly described in a contribution (as it is by Newberry~et~al.) --- but additionally, if the results change given different choices, this should also be reported.
Beyond data operationalisation, we drew attention to issues specific to linguistic data that should be taken into account to ensure quality of testing results, as well as to work in cultural evolution where it has been shown that the inference of individual transmission processes from population-level frequency aggregates is susceptible to error and should be handled with care.  

To conclude, identifying the role and prevalence of stochastic drift in language change is an important goal, but our results suggest that great care should be exercised when applying such tests to linguistic data, in order for the results to not be biased by issues specific to the domain as well as properties of a particular test.

\FloatBarrier

\vspace{10pt}

\section*{Funding information}

The first author of this research was supported by the scholarship program Kristjan Jaak, funded and managed by the Archimedes Foundation in collaboration with the Ministry of Education and Research of Estonia.

\section*{Acknowledgements}

The authors would like to thank Mitchell Newberry for discussion and comments on the paper that led to significant improvements and revisions, the anonymous reviewers for useful comments, and Alison Feder for providing an implementation of the FIT earlier when one was not yet publicly available.

\section*{Competing interests}

The authors declare no competing interests.

\section*{Authors' contributions}

A.K. carried out the research and wrote the paper. R.B., S.K. and K.S. provided revisions, comments and feedback on the design of the research and the paper.

\section*{Data availability}

The R code we used to replicate the results of the original paper is available at \url{https://github.com/mnewberry/ldrift}, and the corpus at \url{https://corpus.byu.edu/coha}. The code to run the simulations described is this paper is available at \url{https://github.com/andreskarjus/wfsim_fit}

\section*{Supplementary appendix}

This appendix expands on the main text, providing additional information, technical details, and further exploration of the parameter spaces of the models.

\subsection*{A note on corpus annotation quality}

While not discussed at length in the main text, the quality of corpus annotation such as lemmatization and part-of-speech tagging plays an equally important role in addition to other corpora-related issues mentioned in the Discussion. Studying the large-scale usage of any linguistic elements of interest relies on the identification of relevant targets in a corpus. Too many erroneously extracted examples can mislead the results. 
Among the 36 verbs in the sample of Newberry~et~al, this is especially pertinent for homonymous words like \textit{wet} and \textit{wed}. 
We already discussed the adjectival usage of \textit{spilt} above. 
We also found that, for example, 44\% of the extracted examples of \textit{wet}.\textsc{past} in the first bin (1812-1875 in COHA, under the variable-width binning procedure) were cases of erroneous tagging --- being instead other non-past forms of \textit{wet} and occurrences of the adjective \textit{wet}. The same issue applies to \textit{wed}, in addition to being confused with the abbreviation for Wednesday.

\subsection*{Results based on a different minimal frequency threshold}

\begin{figure}[thb]
	\renewcommand\thefigure{S1}
	\noindent
	\includegraphics[width=\columnwidth]{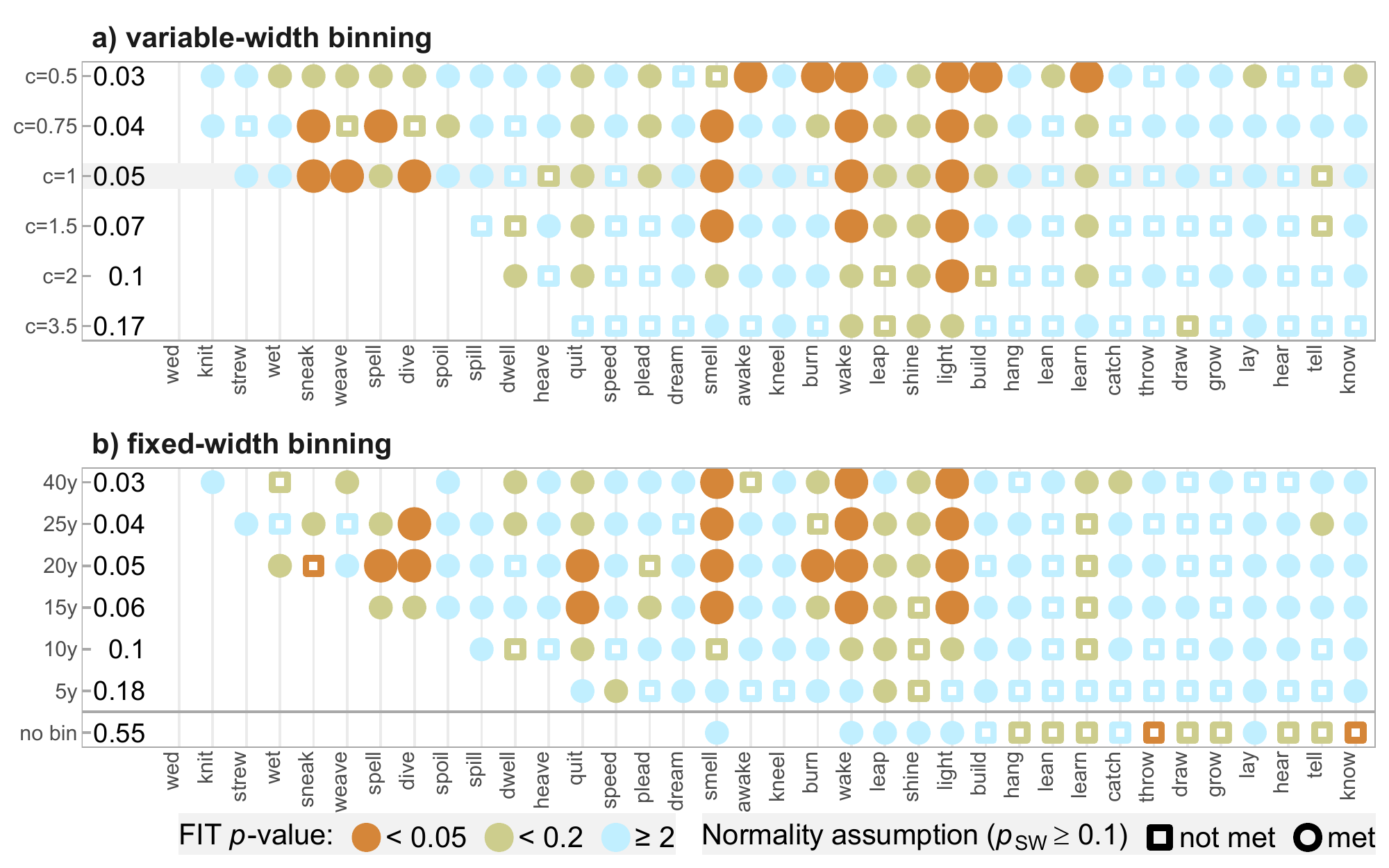}
	\caption{
		Results of applying the FIT to time series constructed based on 200 years of COHA frequency data. 
		The interpretation of this figure is the same as that of Figure~\ref{fig:fitresults}, the only difference being the increased minimal within-bin frequency threshold of 100.
		The constant $c$ determines the number of variable length bins via 
		$n(b) =  c\ln(n(v))$. Thus ``$c=1$'' corresponds to Newberry~et~al.'s original results (highlighted with the horizontal grey line).
		$10{\rm y}$ corresponds to fixed bin length of 10 years, etc; `no bin` refers to no additional binning on top of the default yearly bins in the corpus}
\end{figure}

Figure~S1 is intended to complement Figure~\ref{fig:fitresults}, where we applied a minimal frequency threshold of 10 in each bin (this is mostly relevant for fixed-width binning, as variable-width ensures largely similar bin sizes). Since this is an arbitrary threshold, we also tried a more conservative value of minimal 100 occurrences per bin (for a bin to be included in the time series), with the results reflected in Figure~S1. In summary, the higher threshold does not change the results for variable-width binning, besides some lower-frequency verbs being excluded (the empty lower left corner). In fixed-width binning, some results change, e.g. \textit{spill} is now always flagged as drift, while \textit{burn}, \textit{dive} and \textit{quit} get flagged as selection.

Results of no binning (i.e. using default COHA 1-year bins) should still be taken with a pinch of salt, even when the normality assumption is now met (circles instead of squares) --- removing bins with less than 100 tokens leaves even medium-frequency verbs with only a few bins (e.g., 5 in the case of \textit{light}, spread uneven across 200 years; observe also the median bins-to-years ratio of 0.55).

The fact that the minimal threshold affects fixed binning more is not surprising, as the frequencies vary more. This makes variable-width binning a more attractive solution, but its different behaviour should also be taken into consideration. Should the overall frequency of a pair (or set) of variants change over the course of the corpus, it will end up with more bins over the more frequent end of the time scale. As COHA is not uniform in size across time, having considerably less data per year in the first few decades, time series based on variable-width binning of COHA data systematically have longer segments in the beginning and shorter ones towards the end. The ``long bins" allow for drawing time series over more sparse corpus segments, where fixed binning would yield unreliably small or empty bins. At the same time, variable-width may by nature gloss over some fluctuations (characteristic of drift) while making a series look more smooth (more characteristic of selection).

\FloatBarrier

\subsection*{Results based on series of different lengths}

\begin{figure}[htb]
	\renewcommand\thefigure{S2}
	\noindent
	\includegraphics[width=\columnwidth]{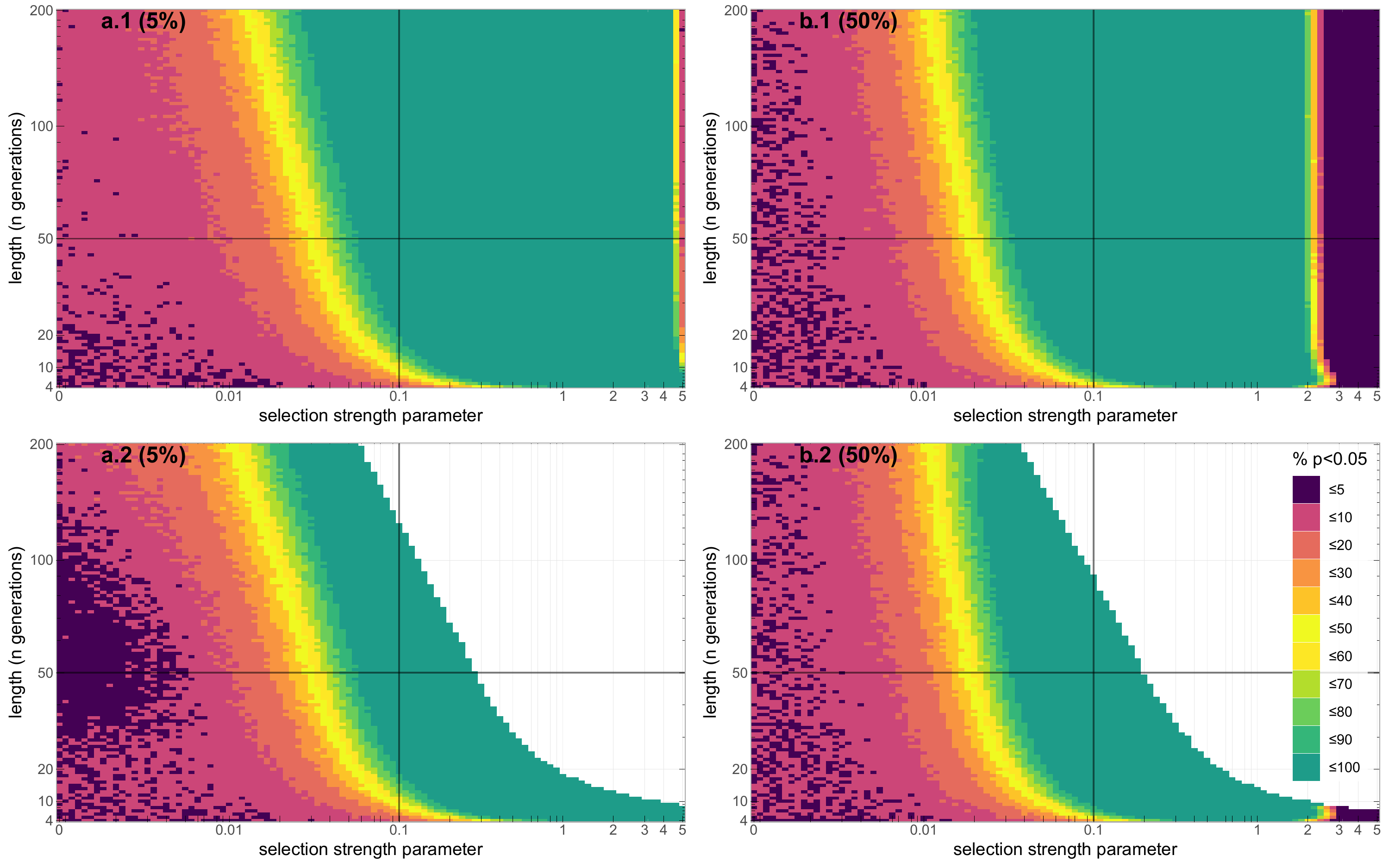}
	\caption{
		The effect of the interplay of time series length and selection strength $s$ on the results of the FIT. The percentage of FIT $p<0.05$ (out of 1000 replications for each combination) is reported for a range of time series lengths (y-axis, $[4,200]$, note the log scale) and the same range of $s$ as above. The left side pair (a) illustrates the case of the time series starting out at 5\%, with the 50\% condition on the right (b). In the bottom panels (a.2, b.2), series with a Shapiro-Wilk $p<0.1$ are removed before calculating the percentage. This figure further illustrates the interplay of series length and $s$ that affect the results of FIT}
\end{figure}

This figure is intended to complement the simulation section in the main text, which focused on the results of binning a 200-length series into shorter series. Here, no binning is being applied. Figure~S2 shows that the FIT produces somewhat different results with the same $s$ given series of different lengths, as expected: when the selection signal is strong enough to be detected (above $\sim 0.02$), then it is easier to detect it in longer series with more data points than in shorter series. Regardless of series length (at least up to the 200), the false positive rate stays in $[0.03,0.07]$ (Figure~S2.b). This also shows that the higher false positive rate under binning shown in Section~\ref{section:wfsimulations} does originate in the binning process (which smooths out small fluctuations) rather than simply length difference (binning naturally also making a series shorter).

\FloatBarrier

\subsection*{More examples of the selection coefficient}

Figure~S3 is intended to complement Figures~\ref{fig:fitmap1}~and~\ref{fig:fitmap3}, where some example Wright-Fisher series where plotted. The $s$ range in our experiments consisted of 200 equidistant values from a log scale between 0.001 and 5, with the addition of 0 in the beginning to be able to explore pure drift.

\begin{figure}[htb]
	\renewcommand\thefigure{S3}
	\noindent
	\includegraphics[width=\columnwidth]{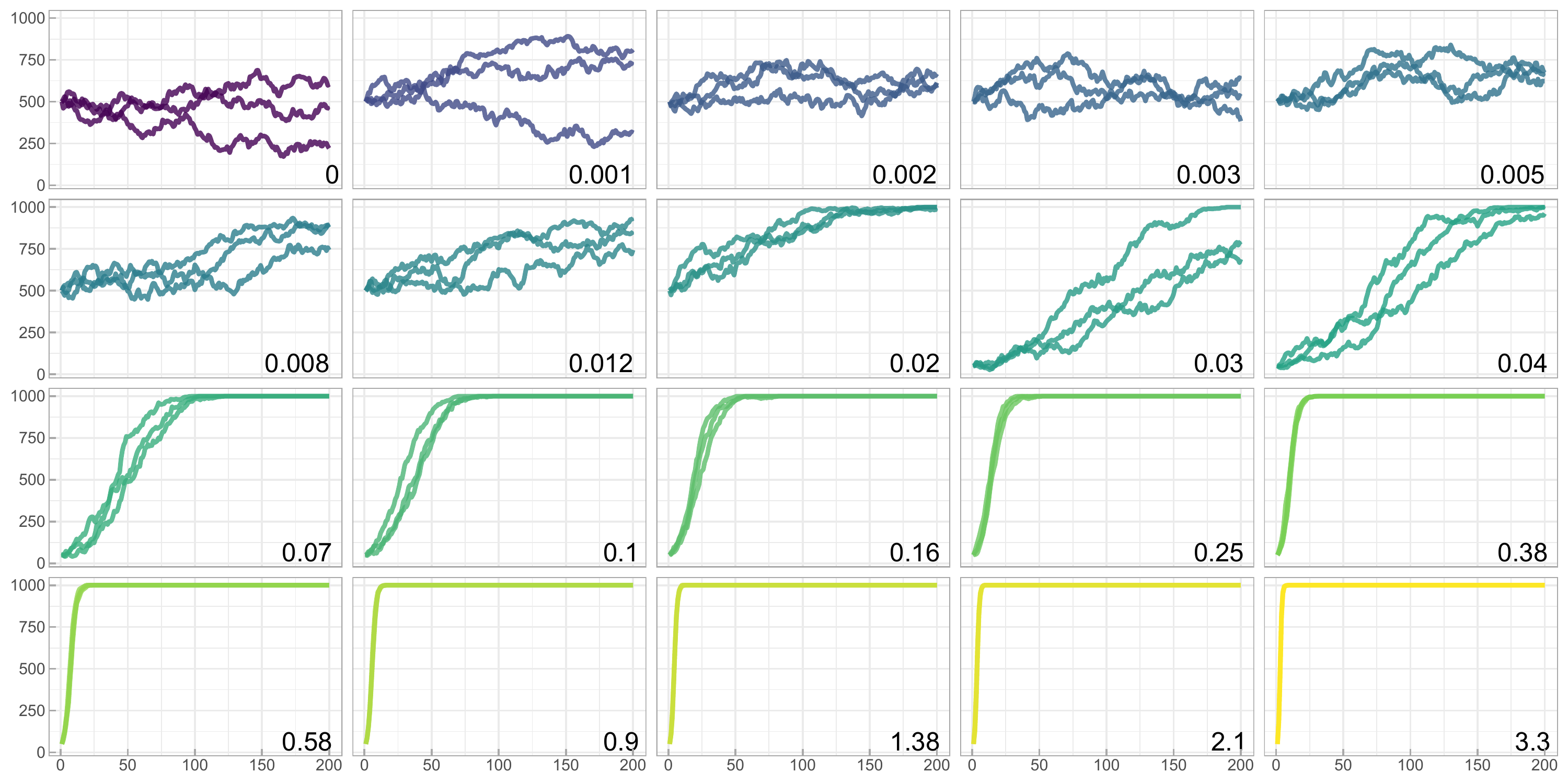}
	\caption{
		A visualization of the range of selection strength $s$ values explored in the simulation section of this study (shown in the corner of each panel). The horizontal axis corresponds to population size (of the `mutant' individuals), with time on the horizontal axis. 
		Higher levels of $s$ lead to the mutants taking over the population at faster rates}
\end{figure}

\FloatBarrier

\subsection*{The increment normality assumption}

The interpretation of the results of the FIT depends how stringently its assumption of the normality of the increments distribution is observed, particularly when $s$ is high.
In Figures~\ref{fig:fitmap2}~and~\ref{fig:fitmap4} we used a Shapiro-Wilk test with a cut-off theshold of 0.1.
We conducted additional simulations to see if a lower would yield qualitatively different results, and found it makes very little difference. We also tried using the Lilliefors-Kolmogorov-Smirnov test and the Anderson-Darling test and found all of them to be broadly in agreement: depending on the series starting point and chosen $\alpha$, the increment normality assumption becomes violated as series (of length 200) approach the $s$ range of $0.05 .. 0.1$, with the breaking point  being somewhat lower on the $s$ scale in non-binned series and higher in series binned into 10-15 bins (i.e. it is easier to meet the normality assumption if the series is binned).

\subsection*{Increment heteroskedasticity and the Fitness Increment Test}

In this additional section, we shed some light on another mathematical aspect of the FIT, the homoskedasticity assumption, as the FIT is, in its core, a one-sample $t$-test for a zero mean under the assumption of normally-distributed increments
with equal variance.
For reference, this is the increment transformation process (cf. Section~\ref{section:fittest}):
\begin{equation}
Y_i = \frac{v_i - v_{i-1}}{\sqrt{ 2v_{i-1}(1-v_{i-1})(t_i-t_{i-1})} }
\end{equation}
where $v_i$ is the relative frequency of a variant in $(0,1)$ at time $t_i$. The rationale behind this rescaling is that, under neutral evolution, the mean increment $v_i-v_{i-1}$ is zero, and its variance is proportional to 
\begin{equation}
v_{i-1}(1-v_{i-1})(t_i-t_{i-1})
\end{equation}
However, here we are dealing with estimates of $v_{i-1}$ and $v_i$ obtained from finite samples of size $M_{i-1}$ and $M_i$, respectively. This leads to additional contributions to the variance of the increment $v_{i}-v_{i-1}$, arising from the variance of the binomial distribution $v(1-v)/M$, where $v$ is the mean value and $M$ is the sample size. To a first approximation, the total variance of the increment is obtained by summing the three contributions. That is, $(v_i - v_{i-1})$ has a variance of 
\begin{equation}
\frac{v_{i-1}(1-v_{i-1})(t_i-t_{i-1})}{N} + \frac{v_{i-1}(1-v_{i-1})}{M_{i-1}} + \frac{v_i(1-v_i)}{M_i} \;.
\end{equation} %
where $N$ stands for effective population size. 
The transform divides all of this by $v_{i-1}(1-v_{i-1})(t_i-t_{i-1})$ which leads to a variance of each \emph{transformed} increment $Y_i$ of approximately
\begin{equation}
\frac{1}{N} + \frac{1}{M_{i-1}(t_i-t_{i-1})} + \frac{v_i(1-v_{i-1})}{ M_i  (t_i-t_{i-1}) v_{i-1}(1-v_{i-1})} \;.
\end{equation}

The FIT can be expected to perform as intended when this variance is constant. This is the case when $1/M \ll (t_i-t_{i-1})/N$ or $M(t_i-t_{i-1}) \gg N$ (assuming $N$ can be inferred, which is not trivial, but cf. Newberry~et~al.). 
The transformed increments based on corpus data basically never have perfectly equal variance once sample size is taken into account, but will be roughly constant when the sample sizes are large (relative to $N$).
The variable-width binning, as employed by Newberry~et~al., assures that the variances are more or less equal, as each bin has roughly the same number of tokens. The worry with fixed-width binning --- including the default data binning of one year in COHA as well further binning of the years into decades and so on --- is that the variance is not going to be equal, as bins may or may not cover a similar number of tokens. 

We calculated these values for the English verb data (with the simplification of excluding $N$, which is not trivial to infer). Variable-width binning consistently yields small increment variances with a very small standard deviation (depending in turn on the variance in the bin sizes in the original data). Using the data without further binning (i.e. 1-year bins from COHA) yields multiple magnitudes higher values for both, as does fixed binning into short bins. But starting at decade-length bins (for higher-frequency verbs like \textit{light}) and 20-year bins (for lower-frequency verbs like \textit{spell}), as bin sizes approach 100 tokens, the picture becomes quite similar to variable-width binning. 

It is not clear, however, how much heteroskedasticity is bad enough to lead to spurious results. For example, is it invalid to interpret the results of the FIT based on 1-year or 5-year bins at all, given typical sample sizes in a corpus like COHA?
While this would benefit from more through future investigation, we attempt to shed some light on this by conducting more Wright-Fisher simulations where we manipulate the size of $M$ in each generation, and the standard deviation of sample sizes ($\sigma_S$), as well as apply different binning strategies to the resulting time series. In Figure~S4, the series length is 200 as in the previous simulations, $s=0$ (as we are interested in the false positives rate), and we explore two $N$ sizes, 10000 (left side column in Figure~S4) and 1000 (right side). Each pixel represents the share of FIT $p<0.05$ in 1000 replications with the given parameter combination.

For each replication in a combination, we run a Wright-Fisher simulation, but to construct the time series, take a random sample of individuals $M$ at each of the 200 generations. The sample sizes are in turn generated by sampling values from a log-normal distribution with a mean of $\ln(M)$ (y-axis in Figure~S4) and $\sigma_S \in \ln([1,2])$ (corresponding to the x-axis in Figure~S4). The log-normal distribution excludes 0, but with a high standard deviation, some of the generated $M$ values can exceed that of the population size, so when reconstructing the time series, they are truncated by taking $\min(M,N)$. After this manipulation however, where $M$ and $V$ are both high, the resulting actual standard deviation across the 200 $M$ sample sizes would not correspond to the predetermined parameter of $\sigma_S$, therefore, such replications are filtered out (along with series where the normality assumption is violated, at Shapiro-Wilk $p<0.1$). When less than 10\% of the replications for a combination are valid, it is excluded from plotting (the white areas in Figure~S4).

The leftmost column of pixels on each panel corresponds to no variance in sample sizes, i.e., the samples are of equal size, corresponding exactly to the value on the y-axis. The top left pixel therefore represents the baseline Wright-Fisher simulation result with no downsampling (the false positive rate being around 5\% in both $N$ at $\alpha=0.05$). Each top panel shows the results without binning, with the lower ones showing results when the series are binned into a smaller number of bins (after the aforementioned downsampling procedure). 

In Figure~S4, where cold blueish colours represent percentages of FIT $p<0.05$. Ideally, as $s=0$, all of the panels should be devoid of any warm colours.
Looking at any single panel, the columns of pixels right of the no-variance leftmost column are not any more yellow than the leftmost column.
This demonstrates that variance in sample sizes does not make any discernable difference --- it does not make the already borderline false positive rate any worse. This observation holds between binning choices. Binning itself does increase the false positive rate, as already determined in Section~\ref{section:wfsimulations} (panels below the top ones exhibit more yellow).
If anything, it would seem series based on samples of size $M<N$ and increased $\sigma_S$ have an improved (i.e. smaller) false positive rate, an effect particularly pronounced when the series are binned. This is however an expected result stemming from the added sampling noise (making any series look more ``random" to the test).

The heteroskedasticity question remains somewhat unresolved, but based on these results we can say that at least the false positive rate of FIT is unlikely to be considerably affected by differing bin sizes. In terms practical guidelines, to be safe, if applicable variable-width binning should be used with FIT as proposed by Newberry~et~al. If fixed-width binning is used, then bins should consist of 100 occurrences or more. 
In the end this is not only a variance problem, but a small sample size problem. A large number of bins consisting each of only tens of occurrences has considerable sampling noise. Given the same corpus, a small number of bins consisting each of hundreds of occurrences can gloss over the true trajectory of change, but also any statistical test based on too few data points is unreliable. In other words, there's no data like more data.

\begin{figure}[htb]
	\renewcommand\thefigure{S3}
	\noindent
	\includegraphics[width=\columnwidth]{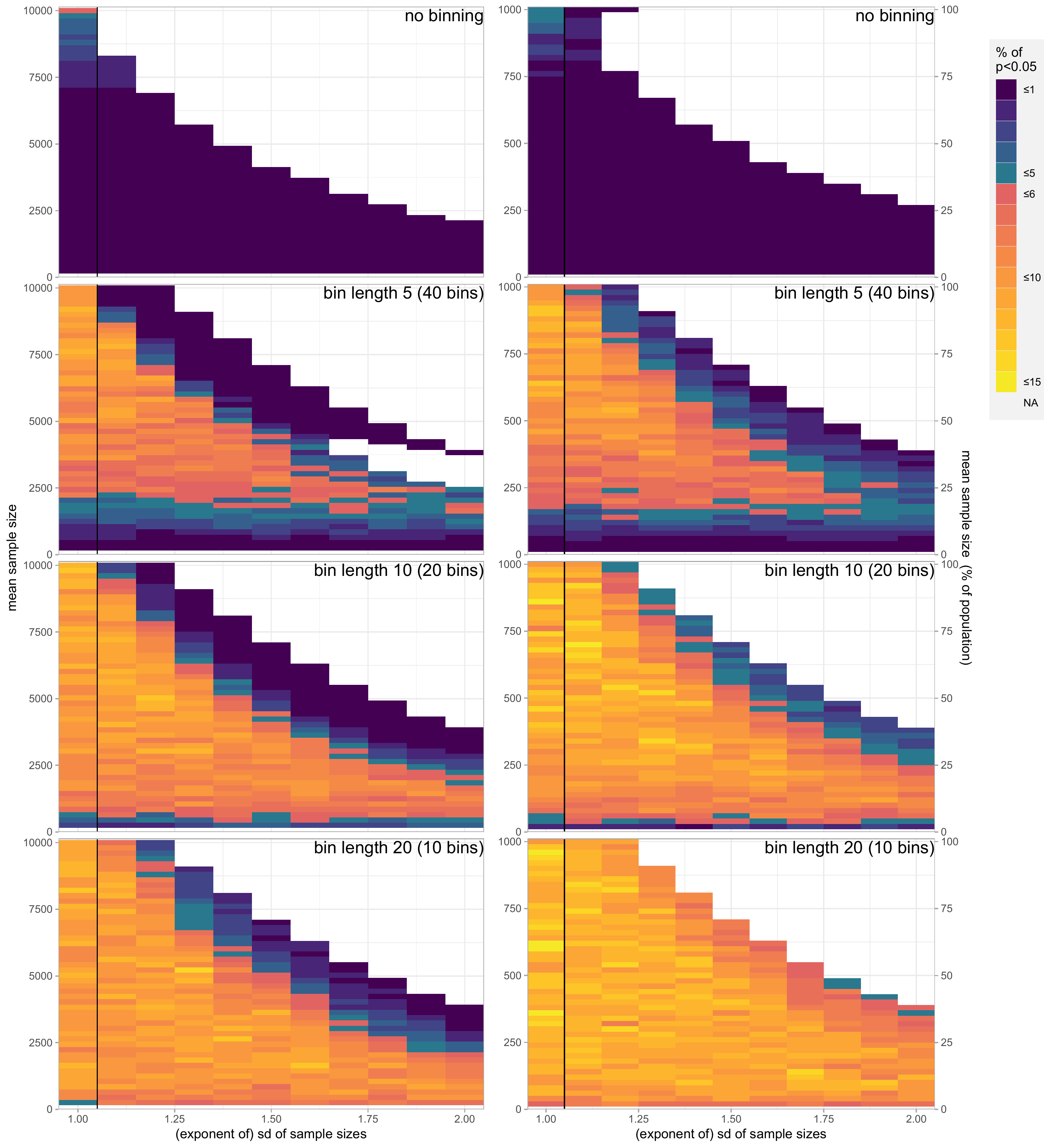}
	\caption{
		False positive rates of the FIT based on Wright-Fisher simulations with downsampled populations.
		Column of panels on the left: $N=10000$. Panels on the right: $N=1000$.
		The cool colours correspond to percentages of $p<0.05$ below 5\%, warm colours indicate higher percentages.
		This figure illustrates that while binning tends to introduce more false positives, in any given binning strategy, added variance in the underlying occurrence counts (and thus bin sizes) does not}
\end{figure}

\FloatBarrier

\printbibliography

\end{document}